\documentclass{article} % For LaTeX2e
\usepackage{iclr2023_conference,times}

\usepackage{hyperref}       % hyperlinks
\usepackage{url}            % simple URL typesetting
\usepackage{booktabs}       % professional-quality tables
\usepackage{xcolor}         % colors

\usepackage{amsmath}
\usepackage{multirow}
\usepackage{amssymb}
\usepackage{bbm}
\usepackage{amsthm}
\usepackage[pdftex]{graphicx}
\usepackage{subcaption}
\usepackage[capitalise]{cleveref}

\theoremstyle{definition}
\newtheorem*{definition}{Definition}

\definecolor{ruby}{HTML}{E02020}
\definecolor{sapphire}{HTML}{0091FF}
\definecolor{amethyst}{HTML}{B620E0}
\definecolor{teal}{HTML}{00A29C}
\definecolor{darkteal}{HTML}{009A94}
\definecolor{blackteal}{HTML}{005A57}
\definecolor{emerald}{HTML}{2EC901}
\definecolor{ioite}{HTML}{6236FF}
\definecolor{citrine}{HTML}{F7B500}
\definecolor{amber}{HTML}{FA6400}
\definecolor{pink}{HTML}{D40272}

\newcommand{\best}[1]{\textbf{#1}}
\newcommand{\nextbest}[1]{\textbf{\textit{#1}}}

\usepackage{amsmath,amsfonts,bm}

\def\eqref#1{equation~\ref{#1}}
\def\1{\bm{1}}

\def\vd{{\bm{d}}}

\def\vk{{\bm{k}}}

\def\vp{{\bm{p}}}
\def\vq{{\bm{q}}}

\def\vv{{\bm{v}}}

\def\vx{{\bm{x}}}
\def\vy{{\bm{y}}}
\def\vz{{\bm{z}}}

\def\mA{{\bm{A}}}

\def\mD{{\bm{D}}}

\def\mK{{\bm{K}}}

\def\mP{{\bm{P}}}
\def\mQ{{\bm{Q}}}

\def\mV{{\bm{V}}}
\def\mW{{\bm{W}}}

\DeclareMathAlphabet{\mathsfit}{\encodingdefault}{\sfdefault}{m}{sl}
\SetMathAlphabet{\mathsfit}{bold}{\encodingdefault}{\sfdefault}{bx}{n}

\newcommand{\vqn}{\vq_\text{new}}
\newcommand{\vkn}{\vk_\text{new}}
\newcommand{\vko}{\vk_\text{old}}

\newcommand{\vvn}{\vv_\text{new}}
\newcommand{\vvo}{\vv_\text{old}}

\DeclareMathOperator*{\Concat}{\mathbin\Vert}

\usepackage[normalem]{ulem}
\usepackage{enumitem}
\setlist[itemize]{topsep=-3pt, itemsep=-3pt, partopsep=1ex, parsep=1ex}

\usepackage{hyperref}
\usepackage{url}

\title{Continual Transformers: Redundancy-Free \\Attention for Online Inference}

\author{Lukas~Hedegaard, Arian~Bakhtiarnia \& Alexandros~Iosifidis 
\\
Department of Electrical and Computer Engineering \\ Aarhus University \\ 
Aarhus, Denmark \\
\texttt{\{lhm,arianbakh,ai\}@ece.au.dk} \\
}

\iclrfinalcopy % Uncomment for camera-ready version, but NOT for submission.
\begin{document}

\maketitle

\begin{abstract}
% Transformers are attention-based sequence transduction models, which have found widespread success in Natural Language Processing and Computer Vision applications.
% Yet, 
Transformers in their common form are inherently limited to operate on whole token sequences rather than on one token at a time.
Consequently, their use during online inference on time-series data entails considerable redundancy due to the overlap in successive token sequences.
In this work, we propose novel formulations of the Scaled Dot-Product Attention, which enable Transformers to perform efficient online token-by-token inference on a continual input stream.
Importantly, our modifications are purely to the order of computations, while the outputs and learned weights are identical to those of the original Transformer Encoder.
We validate our \textit{Continual} Transformer Encoder with experiments on the THUMOS14, TVSeries and GTZAN datasets
with remarkable results:
% To validate our approach, we conduct experiments on visual, audio, and audio-visual classification and detection tasks, i.e. Online Action Detection on THUMOS14 and TVSeries and Online Audio Classification on GTZAN, with remarkable results.
Our \textit{Continual} one- and two-block architectures reduce the floating point operations per prediction by up to 63$\times$ and 2.6$\times$, respectively, while retaining predictive performance.
\end{abstract}

% \IEEEraisesectionheading{\section{Introduction}} \label{sec:introduction}
\section{Introduction} \label{sec:introduction}


% The astounding advance of Deep Neural Networks can be attributed largely to the research community's focus on open source code, common benchmarks, and the continuing endeavour of improving upon prior standards.
%
% \IEEEPARstart{M}{any} 
Many real-life usage scenarios such as the perception in self-driving cars
% , agile production with human-robot interaction, 
and live monitoring of critical resources process a continual stream of inputs and require near-instantaneous predictions per time-step.
This stands in contrast to what many common benchmarks for deep learning evaluate, namely the operation on distinct batches of data with no inter-batch relationships.
Consequently, a plethora of methods have been developed~\citep{ju2013conv, carreira2017quo, varol2018long, yan2018spatial, heidari2021progressive, vaswani2017attention, Arnab2021ViViTAV, 2106.15183}, which focus on batch-wise processing, but fail to optimise for online operation, where new information (e.g., a video frame / token) arrives at each step from a continual input stream, and future information is not available at the current time-step.
% Examples of this include 3D Convolutional Neural Networks (CNNs)~\citep{ju2013conv, carreira2017quo, varol2018long}, Spatio-Temporal Graph Neural Networks (ST-GCNs)~\citep{yan2018spatial, heidari2021progressive} and Transformer style-models~\citep{vaswani2017attention, Arnab2021ViViTAV, 2106.15183}. 
We need a class of networks, which operate efficiently on\textit{ both batches of data and on continual streams}. 
% Continual Inference Networks (CINs, \cref{sec:cin}) serve this purpose. 

%Continuing the line of work on \textit{Continual} 3D CNNs~\citep{hedegaard2021co3d} and \textit{Continual} ST-GCNs~\citep{hedegaard2022online},
% Continuing the line of work on CINs~\citep{hedegaard2021co3d, hedegaard2022online, hedegaard2022colib}, 
Accordingly, we propose a reformulation of the Transformer Encoder as a Continual Inference Network (CIN, \cref{sec:cin}) which accelerates the stream processing on time-series data, while retaining weight-compatibility.
Specifically, we derive two variants of Continual Scaled Dot-Product Attention (SDA) for the cases where prior output tokes \textit{should}
% (see \cref{fig:co-re-dot-prod-attention}) 
and \textit{should not} 
% (see \cref{fig:cosi-dot-prod-attention}) 
be updated after observing a new input token.
Notably, our attention formulations reduce the per-step cost of SDA~\citep{vaswani2017attention} from time complexity
$\mathcal{O}({n^2d})$ to $\mathcal{O}({nd})$ and memory complexity $\mathcal{O}({n^2})$ to $\mathcal{O}({nd})$ and are readily embedded into Continual Multi-Head Attention (MHA) and Continual Transformer Encoder blocks. 
Finally, we propose the use of Recycling Positional Encoding to accommodate progressive caching of partial attention results for continual data streams.
% To inject positional information in a continual stream of inputs, we further propose Recycling Positional Encoding, which add positional information in a round-robin schedule to accommodate progressive caching of prior partial attention results. 

Due to the interdependence of SDA outputs, Continual Transformers are most efficient for shallow architectures. Shallow Transformers have many applications such as augmentations of CNNs~\citep{touvron2021augmenting}, light-weight Natural Language Processing~\citep{cornia2020smart}, fusion operations in multi-modal (e.g. audio-visual) settings~\citep{chumachenko2022self} and early exit branches in multi-exit architectures~\citep{2105.09121, 2106.15183}.
% Though our innovation works best for shallow Transformers, these have diverse applicability,
% for instance in augmenting CNNs~\citep{touvron2021augmenting}, in structured content extraction~\citep{shen2022vila}, and for early exit architectures~\citep{2105.09121}.
In our experiments\footnote{
% Our source code is available at \url{https://github.com/lukashedegaard/continual-transformers}.
Source is code provided in supplementary material. Link will be made available upon acceptance.
}, we validate their exceptional efficiency improvements on common benchmarks in Online Action Detection~\citep{idrees2017thumos} and Online Audio Classification~\citep{tzanetakis2001gtzan}. 

\begin{figure}[tb]
    \centering
    \includegraphics[width=\linewidth]{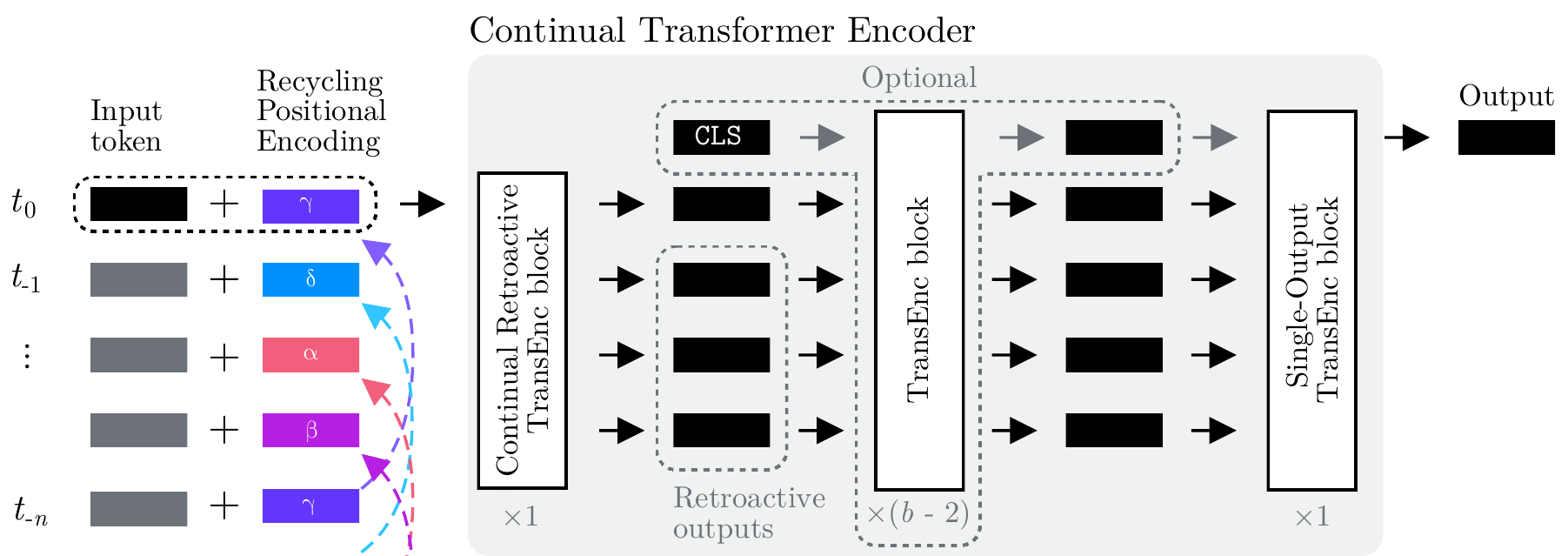}
    \caption{
        \textbf{Multi-block Continual Transformer Encoder with Recycling Positional Encoding.} For $b>2$ blocks, regular Transformer Encoder blocks can be added between an initial Continual Retroactive block and a final Single-Output block. A class-token may be used after the initial block.
    }
    \label{fig:three-layer-architecture}
\end{figure}
\section{Related Work}

\subsection{Continual Inference Networks} \label{sec:cin}

\begin{definition}[\textbf{Continual Inference Network}]
\itshape
A Deep Neural Network, which
\begin{itemize}%[\leavevmode\vspace{-0.4\baselineskip}]%[wide=0.5em, leftmargin =*, nosep, before = \leavevmode\vspace{-\baselineskip}]
    \item is capable of continual step inference without computational redundancy,
    \item is capable of batch inference corresponding to a non-continual Neural Network,
    \item produces identical outputs for batch- and step inference given identical receptive fields,
    \item uses one set of trainable parameters for both batch and step inference.
\end{itemize}
\end{definition}

\noindent These requirements ensure that a Neural Network has broad applicability for both (offline) batch-wise inference (i.e., most research benchmarks) and online stream processing. 
% Since non-CINs can be trivially applied to process data streams by keeping a cache of prior step-inputs and aggregating 
% An important distinction between CINs and non-CINs, is the ability to perform redundancy-free step-wise inference. 
While non-CINs can operate on streams of data by caching prior steps in a first-in first-out (FIFO) queue and aggregating them to a full \mbox{(spatio-)}temporal input, which is processed similarly to an offline batch, this entails computational redundancy in proportion with the sequence length. %, as illustrated in \cref{fig:redundancy}.
CINs perform step-wise inference without such caching and repeat computation.
Uni-directional Recurrent Neural Networks are an example of Continual Inference Networks. Their default mode of operation is by time-step and they are easily applied to spatio-temporal batches of data by concatenation of the step-wise outputs. 
% Given their ability to operate efficiently on either batches or continual input streams, RNNs may be considered the first commonly used Continual Inference Networks.
%
Recently, a modification to the spatio-temporal 3D convolution was proposed~\citep{hedegaard2021co3d}, which enables existing 3D CNNs to operate efficiently during continual inference. A similar principle was used to enhance Spatio-temporal Graph Convolutions as well~\citep{hedegaard2022online}. %by weights transfer to a \textit{Continual} 3D CNN. Importantly, \textit{Co}3D CNNs produce identical output to that of regular 3D CNNs during regular step-wise inference, and the learned weights are directly transferable between regular 3D CNNs and \textit{Co}3D CNNs. This enables impressive speed-ups during online operation of 3D CNN, which would otherwise entail considerable computational redundancy.
%
% A similar redundancy is experienced by Transformer models applied to continual time-series processing. In this work, we continue the effort of converting existing state-of-the-art architectures to Continual Inference Networks and propose alternative computational schemes for the Scaled Dot-Product Attention operation, which enable efficient operation of shallow Transformers on continual input streams.
In \cref{sec:cotrans}, we derive a CIN formulation for Transformer Encoders.

% \begin{figure}[t]
%     \centering
%     \includegraphics[width=0.6\linewidth]{figures/RedundancyIllustration-alt-colors.pdf}
%     \caption{
%         \textbf{Redundant computations during online inference} for a regular two-layer neural network with temporal connectivity and receptive field of three.
%         Contributing connections for the \textcolor{amber}{prior step $t_{-1}$} and \textcolor{ioite}{current step $t_{0}$} alongside \mbox{\textcolor{pink}{redundant}} computations are highlighted.
%     }
%     \label{fig:redundancy}
% \end{figure}

\subsection{Transformer architectures} \label{sec:trans}

Initially proposed for sequence-to-sequence modelling in Natural Language Processing, the Transformer~\citep{vaswani2017attention} has become a canonical building block in many applications of Deep Learning, 
including Computer Vision~\citep{dosovitskiy2021an, Arnab2021ViViTAV, wang2021oadtr, carion2020end} and Audio Classification \citep{gong21b_interspeech}.
% including Computer Vision tasks such as Image Classification~\citep{dosovitskiy2021an}, Video Recognition~\citep{Arnab2021ViViTAV, wang2021oadtr}, and Object Detection~\citep{carion2020end} as well as Audio Classification \citep{gong21b_interspeech}.
%
Their success can be partly attributed to reduced inductive bias compared with CNNs and RNNs, which allows better adaptations when sufficiently large datasets are available; 
the Scaled Dot-Product Attention (SDA) maps a set of input tokens to a set of outputs without inherent preconceptions.
% However, this flexible many-to-many attention has a time and space complexity, which exhibits quadratic growth with the number of tokens in the set.
However, this many-to-many attention exhibits quadratic growth in time and space complexity with the token count $n$ in the set.

A great deal of research has sought to improve the efficiency of Transformers~\citep{2009.06732}. Block-wise or Chunking methods such as Image Transformer~\citep{pmlr-v80-parmar18a} and Vision Transformer~\citep{dosovitskiy2021an} group up entities of a local receptive field into a single block, reducing the $ \mathcal{O}(n^2) $ complexity to $ \mathcal{O}(n_b^2) $, where $ n_b < n $ is the number of blocks. Techniques such as sliding windows, dilation and pooling can be used to achieve a similar effect~\citep{beltagy2020longformer}. 
% As opposed to a fixed grouping scheme, 
The Reformer~\citep{kitaev2020reformer} reduces the complexity to $\mathcal{O}(n \log n)$ by learning groupings in a data-driven manner via Locality-Sensitive Hashing (LSH).
% More recently, a group of attention-free Multi-Layer Perceptron (MLP) based approaches such as MLP-Mixer~\citep{2105.01601} and ResMLP~\citep{2105.03404} have been proposed, that strive to obtain performance similar to that of Transformers, while reducing the computational cost by removing the Self-Attention mechanism all together and employing MLPs in conjunction with transposition in order to preserve a global receptive field~\citep{Guo2021}.
%
A different paradigm aims to derive approximations of the self-attention matrix. Methods such as Linformer~\citep{2006.04768}, Nyströmformer~\citep{xiong2021nystromformer} and Performer~\citep{choromanski2021rethinking} reduce the complexity from $\mathcal{O}(n^2)$ to $\mathcal{O}(n)$. 
Unlike these efforts, our approach produces the \textit{exact} same computational outputs for temporal sequences as the original Multi-Head Attention. % and retains full weight-compatibility.
\section{Continual Transformers} \label{sec:cotrans}
In this work, we examine the use of Transformer Encoders for stream-processing, where we receive one token per time-step.
Specifically, the query, key and value inputs constitute a continual stream of $d$-dimensional tokens and we wish to compute the outputs for each step immediately considering $n-1$ prior tokens. 
We begin our exposition in \cref{sec:reg-attn} by considering the Scaled Dot-Product Attention (SDA) for this task. 
To alleviate the inefficiencies of SDA, we propose two alternative computational sequences in \cref{sec:co-re-attn} and \cref{sec:co-si-attn} and compare them to SDA in \cref{sec:co-comparison}. Finally, sections \ref{sec:co-mha}-\ref{sec:arch-considerations} build up the full architecture, and discuss architectural considerations.
% First, let us examine three SDA implementations and derive the complexity of each.

\subsection{Regular Scaled Dot-Product Attention} \label{sec:reg-attn}
% Using the symbolism from https://arxiv.org/pdf/2009.14794.pdf
% Here, the softmax is broken out with D reprensenting the denominator of the softmax-fraction.
Denoting query, key, and value sequence matrices by $\mQ, \mK, \mV \in \mathbbm{R}^{n \times d}$, 
the regular Scaled Dot-Product Attention first defined by \citet{vaswani2017attention} can be written as:
\noindent
\begin{equation}
    \text{Att}(\mQ, \mK, \mV) = \mD^{-1} \mA \mV
    \qquad
    \mA = \text{exp}\left(\mQ \mK^{\top} / \sqrt{d} \right)
    \qquad
    \mD = \text{diag}\left( \mA \mathbbm{1}_n ^\top \right),
        \label{eq:scaled-dot-product-attention}
\end{equation}
%
% \noindent
% \begin{minipage}{.32\linewidth}
%     \begin{equation}
%         \text{Att}(\mQ, \mK, \mV) = \mD^{-1} \mA \mV  
%         \label{eq:scaled-dot-product-attention-att}
%     \end{equation}
% \end{minipage}%
% \begin{minipage}{.32\linewidth}
%     \begin{equation}
%         \mA = \text{exp}\left(\mQ \mK^{\top} / \sqrt{d} \right)            
%         \label{eq:scaled-dot-product-attention-a}
%     \end{equation}
% \end{minipage}
% \begin{minipage}{.32\linewidth}
%     \begin{equation}
%     \mD = \text{diag}\left( \mA \mathbbm{1}_n ^\top \right),
%     \label{eq:scaled-dot-product-attention-d}
%     \end{equation}
% \end{minipage}
%
% \begin{align}
%     \text{Att}(\mQ, \mK, \mV) &= \mD^{-1} \mA \mV 
%         \label{eq:scaled-dot-product-attention-att} \\
%     \mA &= \text{exp}\left(\mQ \mK^{\top} / \sqrt{d} \right)            
%         \label{eq:scaled-dot-product-attention-a} \\
%     \mD &= \text{diag}\left( \mA \mathbbm{1}_n ^\top \right),
%         \label{eq:scaled-dot-product-attention-d}
% \end{align}
%
% where $\mQ, \mK, \mV \in \mathbbm{R}^{n \times d}$ are query, key, and value row-matrices, 
where $\mA, \mD \in \mathbbm{R}^{n \times n}$ and $\mathbbm{1}_n$ is a row-vector of $n$ ones.
In each time-step, we can update $\mQ$, $\mK$, and $\mV$ by discarding their oldest token and prepending a new one in a FIFO manner. 
This is a common implementation for step-wise inference, e.g. found in the \textsc{fairseq} library~\citep{ott2019fairseq}.
% Then the computational steps in \cref{eq:scaled-dot-product-attention} are repeated again.

Each time-step results in $2n^2d+2nd$ multiplications, $2n^2d - nd - n$ additions, and $n^2$ exponentiations as accounted for in \cref{apx:mha-scaling}, which amounts to a time complexity of $\mathcal{O}({n^2d})$ and a $\mathcal{O}({n^2})$ memory complexity originating from the transient feature-map $\mA$. Furthermore, a constant-sized cache of size $3(n-1)d$ is needed to store the $n-1$ latest tokens in $\mQ$, $\mK$ and $\mV$.
We could avoid considerable redundancy by caching $\mQ \mK^{\top}$ directly. However, this comes with a memory penalty of $(n-1)^2$. 
Fortunately, another computational scheme can be devised. %, which is laid out in \cref{sec:co-re-attn}.

% \begin{table}[!htbp]
% \caption{
%     \textbf{Floating Point Operations} for the Scaled Dot-Product Attention in
%     \cref{eq:scaled-dot-product-attention}. $\mD^{-1} (\cdot)$ can be efficiently computed as element-wise multiplication with $\mA \mV$.
% }\label{tab:dot-prod-complexity}
% \begin{center}
% % \resizebox{\linewidth}{!}{
% \begin{tabular}{l|ccc}
%     \toprule
%         & Mul.              & Add               & Exp       \\
%     \midrule
%     Eq. (\ref{eq:scaled-dot-product-attention}.1) 
%         & $n^2d + nd$       & $nd(n-1)$      & 0 \\
%     % & $\mA \mV$           & $n^2d$        & $n^2(d-1)$        &           \\
%     % & $\mD^{-1} (\cdot)$  & $n+nd$          &                   &           \\
%     Eq. (\ref{eq:scaled-dot-product-attention}.2) 
%         & $n^2d + nd$       & $n^2(d-1)$    & $n^2$ \\
%     % & $\mK / \sqrt{d}$    & $nd$          &                   &           \\
%     % & $\mQ \mK^{\top}$    & $n^2d$        & $n^2(d-1)$        &           \\
%     % & $exp(\cdot)$        &               &                   & $n^2$     \\
%     Eq. (\ref{eq:scaled-dot-product-attention}.3) 
%         & $0$               & $n(n-1)$      &  0 \\
%     \bottomrule
% \end{tabular}
% % }
% \end{center}
% \end{table}

% \newpage
\subsection{Continual Retroactive Scaled Dot-Product Attention} \label{sec:co-re-attn}

We can compute $\mD^{-1}\mA\mV$ in a step-wise manner using the latest query, key, and value steps, $\vqn, \vkn, \vvn \in \mathbbm{R}^{1 \times d}$, alongside appropriately cached partial results.
The softmax normalisation with $\mD^{-1}$ can be efficiently implemented via column-aligned element-wise multiplications (denoted by $\odot$ hereafter) of a column-vector $\vd = \mA\mathbbm{1}_n^\top$.
If we cache the $n-1$ values for the prior step tokens, i.e. \mbox{$\vd_{\text{mem}} = \mA^{(-n+1:-1)}_\text{prev}\mathbbm{1}_{n-1}^\top$}, alongside $\mQ$ and $\mK$, we can define the step update as:
\begin{align}
    \vd^{(-n+1:-1)} 
        &= \vd_\text{mem}^{(-n+2:0)} 
        - \text{exp}\left( \mQ_\text{mem}\vko^\top \right)
        + \text{exp}\left( \mQ_\text{mem}\vkn^\top\right) 
    \label{eq:core-sdpa-d-update}
    \\
    \vd^{(0)} &= \text{exp}\left( \frac{\vqn}{\sqrt{d}}\left(\mK_\text{mem} \mathbin\Vert  \vkn \right)^\top \right) \mathbbm{1}_n^\top,
    \label{eq:core-sdpa-d-0}
\end{align}
%
% \begin{equation}
%     \vd^{(0)} = \text{exp}\left( \frac{\vqn}{\sqrt{d}}\left(\mK_\text{mem} \mathbin\Vert  \vkn \right)^\top \right) \mathbbm{1}_n^\top,
%     \label{eq:core-sdpa-d-0}
% \end{equation}
%
where $\mQ_\text{mem}$ ($\mK_\text{mem}$) are the $n-1$ prior query (key) tokens, $\vko$ is the key from $n$ steps ago, and $\mathbin\Vert$ denotes concatenation of matrices along the first dimension.
Negative indices indicate prior time-steps.
An update for $\mA\mV$ can likewise be defined as a function of the $n-1$ prior values $\mA\mV_\text{mem}$:
\begin{align}
    \mA\mV^{(-n+1:-1)} &= 
        \mA\mV_\text{mem}^{(-n+2:0)} 
        - \text{exp}\left(\mQ_\text{mem} \vko^\top \right) \vvo  
        + \text{exp}\left(\mQ_\text{mem} \vkn^\top \right) \vvn 
    \label{eq:core-sdpa-av-update}
    \\
    \mA\mV^{(0)} 
        &= \text{exp}\left(
            \frac{\vqn}{\sqrt{d}} 
            \left(\mK_\text{mem} \mathbin\Vert  \vkn \right)^\top
        \right)
        \left(\mV_\text{mem} \mathbin\Vert  \vvn \right).
    \label{eq:core-sdpa-av-0}
\end{align}
%
% \begin{equation}
%     % \\[6pt]
%     % \mA\mV^{(-n+1:-1)} &= \mA\mV_\text{mem}^{(-n+2:0)} + \mQ_\text{mem} \left(\vkn^\top \vvn - \vko^\top \vvo \right) \\
%     %
%     \mA\mV^{(0)} 
%         % &= \frac{\vqn}{\sqrt{d}} \vkn^\top \vvn + \frac{\vqn}{\sqrt{d}} \mK_\text{mem}^\top \mV_\text{mem}
%         = \text{exp}\left(
%             \frac{\vqn}{\sqrt{d}} 
%             \left(\mK_\text{mem} \mathbin\Vert  \vkn \right)^\top
%         \right)
%         \left(\mV_\text{mem} \mathbin\Vert  \vvn \right)
%     \label{eq:core-sdpa-av-0}
% \end{equation}
%
Finally, we compute the Continual Retroactive Attention output in the usual manner:
\begin{equation}
    CoRe\text{Att}(\vqn, \vkn, \vvn) 
    % = \text{diag}(\vd)^{-1} \mA \mV.
    = \vd^{-1} \odot \mA \mV.
    \label{eq:core-sdpa-dav}
\end{equation}
An visual depiction of these update steps is provided in \cref{apx:sup-vis}. %\cref{fig:co-re-dot-prod-attention}.
% As to the computational complexity, 
A time-step can now be computed with $7nd + 2n - 3d$ multiplications, $6nd + 3n - 6d -3$ additions, and $3n - 2$ exponentials. % (details found in \cref{tab:core-dot-prod-complexity}).
This time complexity of $\mathcal{O}({nd})$ per step and a $\mathcal{O}({nd})$ memory complexity is a significant improvement over the prior $\mathcal{O}({n^2d})$ and $\mathcal{O}({n^2})$ complexities in \cref{sec:reg-attn}.

% \begin{table}[!htbp]
% \caption{
%     \textbf{Floating Point Operations} for the Continual Retroactive Dot-Product Attention in
%     \crefrange{eq:core-sdpa-d-update}{eq:core-sdpa-dav}.
%     The outputs of the exponentials in \cref{eq:core-sdpa-d-update} and \cref{eq:core-sdpa-d-0} can be reused in \cref{eq:core-sdpa-av-update} and \cref{eq:core-sdpa-av-0} respectively, and are omitted in the count.
% }\label{tab:core-dot-prod-complexity}
% \begin{center}
% % \resizebox{\linewidth}{!}{
% \begin{tabular}{l|ccc}
%     \toprule
%     &           Mul.            & Add                   & Exp       \\
%     \midrule
%     \cref{eq:core-sdpa-d-update} 
%                 & $2(n-1)d$     & $2(n-2)d + 2(n-1)$    & $2(n-1)$  \\
%     \cref{eq:core-sdpa-d-0} 
%                 & $nd + n + d$  & $nd + (n-1) + d$      & $n$       \\
%     \cref{eq:core-sdpa-av-update} 
%                 & $2(n-1)d$     & $2(n-1)d$             & 0         \\
%     \cref{eq:core-sdpa-av-0} 
%                 & $nd$          & $(n-1)d$              & 0         \\
%     \cref{eq:core-sdpa-dav} 
%                 & $nd+n$        & 0                     & 0         \\

%     \bottomrule
% \end{tabular}
% % }
% \end{center}
% \end{table}

% \newpage
\subsection{Continual Single-Output Scaled Dot-Product Attention} \label{sec:co-si-attn}

% \begin{figure}[b]
%     \centering
%     \includegraphics[width=0.6\linewidth]{figures/CoStepDotProductAttention.pdf}
%     \caption{
%         \textbf{Continual Single-Output Dot-Product Attention}. 
%         The key ($\mK$) and value ($\mV$) matrices are aggregated over time by caching the step vectors $\vkn$ and $\vvn$ in a FIFO queue. During each step, only the attention output associated with $\vq$ is computed.
%     }
%     \label{fig:cosi-dot-prod-attention}
% \end{figure}

% In the context of time-series processing, 
Both the Regular and Continual Retroactive Dot-Product Attentions produce attention outputs for the current step, as well as $n-1$ retroactively updated steps.
In cases where retroactive updates are not needed, we can simplify the computation greatly via a Continual Single-Output Dot-Product Attention ($CoSi\text{Att}$). % as depicted in \cref{fig:cosi-dot-prod-attention}.
In essence, the regular SDA is reused, but prior values of $\mathbf{k}$ and $\mathbf{v}$ are cached between steps (as in \citep{ott2019fairseq}), and only the attention corresponding to a single query token $\vq$ is computed:
\begin{equation}
    CoSi\text{Att}(\vq, \vkn, \vvn) 
        = \mathbf{a} \left(\mV_\text{mem} \mathbin\Vert  \vvn \right) 
        / \mathbf{a}\mathbbm{1}_n^\top
        \label{eq:leading-scaled-dot-product-attention-att}, \qquad
    \mathbf{a} = \text{exp}\left(
        \frac{\vq}{\sqrt{d}} 
        % \left(\vq / \sqrt{d} \right)
        \left(\mK_\text{mem} \mathbin\Vert  \vkn \right)^{\top} 
    \right).
        % \label{eq:leading-scaled-dot-product-attention-a}
    % \mK = \left(\mK_\text{mem} \mathbin\Vert  \vkn \right)
    %     \label{eq:leading-scaled-dot-product-attention-k}
\end{equation}

% An illustration is provided in \cref{apx:sup-vis}
A step output is computed with $2nd + 2d$ multiplications, $2nd - d - 1$ additions, and $n$ exponentials. The time- and memory complexities remain $\mathcal{O}({nd})$ per step. % and the memory complexity is $\mathcal{O}({nd})$ as seen in \cref{tab:resi-dot-prod-complexity}. 
Using the (leading) query $\vqn$ as input, the attention is purely causal. % in that the output is based only on previously seen values. 
Alternatively, prior (lagging) query vectors could be cached and used as query input, though this would introduce a network delay.
%
% A (non-continual) Single-Output SDA, $Si\text{Att}(\vq, \mK, \mV)$, also has applications during batch inference. 
% Here, the cached key and value steps are simply replaced by the original key and query matrices, and the attention output corresponding to a single query token (e.g. task token) can be computed.

% \begin{table}[!htbp]
% \caption{
%     \textbf{Floating Point Operations} for the Continual Single-Output SDA in \cref{eq:leading-scaled-dot-product-attention-att}.
% }\label{tab:resi-dot-prod-complexity}
% \begin{center}
% % \resizebox{\linewidth}{!}{
% \begin{tabular}{l|ccc}
%     \toprule
%     &           Mul.            & Add                   & Exp       \\
%     \midrule
%     Eq. (\ref{eq:leading-scaled-dot-product-attention-att}.1) 
%                 & $nd + d$     & $(n-1)d + n - 1$       & $0$  \\
%     Eq. (\ref{eq:leading-scaled-dot-product-attention-att}.2) 
%                 & $nd + d$  &   $n(d-1)$                & $n$       \\
%     \bottomrule
% \end{tabular}
% % }
% \end{center}
% \end{table}

% \newpage
\subsection{Comparison of Scaled Dot-Product Attentions} \label{sec:co-comparison}
% Let us briefly compare the three attention types.
Assuming $n-1$ prior $\vq$, $\vk$ and $\vv$ steps have been calculated by the Continual SDA modules, and that 
\mbox{$\mQ = \left(\mQ_\text{mem} \mathbin\Vert  \vqn \right)$}, 
\mbox{$\mK = \left(\mK_\text{mem} \mathbin\Vert  \vkn \right)$}, and 
\mbox{$\mV = \left(\mV_\text{mem} \mathbin\Vert  \vvn \right)$},
we have the correspondence:% between the Scaled Dot-Product Attention types
%
% \begin{align}
%     \text{Att}(\mQ, \mK, \mV)^{(t)}
%     &= CoRe\text{Att}(\vqn, \vkn, \vvn)^{(t)} 
%     \nonumber \\ &= CoSi\text{Att}(\mathbf{q}_t, \vkn, \vvn)
%     \nonumber \\ &= Si\text{Att}(\mathbf{q}_t, \mK, \mV).
%     \label{eq:co-correspondence}
% \end{align}
%
\begin{equation}
    \text{Att}(\mQ, \mK, \mV)^{(t)}
    = CoRe\text{Att}(\vqn, \vkn, \vvn)^{(t)} 
    = CoSi\text{Att}(\mathbf{q}_t, \vkn, \vvn)
    % = Si\text{Att}(\mathbf{q}_t, \mK, \mV).
    \label{eq:co-correspondence}
\end{equation}
Here, $\vq_t$ is the $t^\text{th}$ row of $\mQ$, i.e. $\mQ^{(t)}$.
During stream processing, the complexity of the Continual Retroactive SDA scales significantly more favourably that the regular SDA. 
For example, the floating point operations (FLOPs) are reduced by $31\times$ when $n=d=100$ and $308\times$ when $n=d=1000$. 
If retroactive output updates are not needed, the Continual Single-Output SDA reduces FLOPs by respectively $100\times$ and $1000\times$.
% $\approx3.25\times$%\footnote{For $d \rightarrow \infty, n \rightarrow \infty$ while counting FLOPs from multiplications, additions and exponentiations equally.}) 
% with the Continual Single-Output SDA. 
The scaling properties are detailed in \cref{apx:mha-scaling}.
\subsection{Continual Multi-Head Attention} \label{sec:co-mha}
Continual Scaled Dot-Product Attentions can replace regular SDA's directly in a Multi-Head Attention (MHA).
Given a new query, key, and value, $\vq, \vk, \vv$, the Continual MHA is defined as
\begin{align}
    Co\text{MHA}(\vq, \vk, \vv) 
    = \left(\Concat_{i=0}^{h-1} Co\text{Att}(
        \vq \mW_Q^{i}, \vk \mW_K^{i}, \vv \mW_V^{i}
    )\right) \mW_O,
\end{align}
where $\Concat$ denotes concatenation of $h$ heads and $\mW_Q^{i}, \mW_K^{i} \in \mathbbm{R}^{d \times d_K/h}$, $\mW_V^{i} \in \mathbbm{R}^{d \times d_V/h}$, and $\mW_O \in \mathbbm{R}^{d_V \times d_O}$ are projection matrices
of head $i$.
% for query, key, and value of head $i$ as well as output.
$Co\text{Att}$ can be either 
$CoRe\text{Att}$ or $CoSi\text{Att}$. 
%the retroactive of single-output SDA.

\subsection{Continual Transformer Encoder} \label{sec:co-trans-enc}
A Continual MHA block can be integrated in a Continual Transformer Encoder block as follows:
% Given a row-vector $\vx$ corresponding to the newest step input, an encoder block follows the two-step process:
%
% \begin{align}
%     \vy &= \text{LayerNorm} \left(\text{Sel}(\vx) + Co\text{MHA}(\vx, \vx, \vx) \right) \\
%     \vz &= \text{LayerNorm} \left(\vy + \text{FF}(\vy) \right),
% \end{align}
%
\begin{equation}
    \vz = \text{LayerNorm} \left(\vy + \text{FF}(\vy) \right),
    \qquad
    \vy = \text{LayerNorm} \left(\text{Sel}(\vx) + Co\text{MHA}(\vx, \vx, \vx) \right),
\end{equation}
where $\vx$ corresponds to the newest step input and $\text{Sel}(\cdot)$ selects a single (last) token of $\vx$ if \textit{CoSi}MHA is used, or selects all tokens otherwise. 
$\text{FF}(\cdot)$ is a two-layer feed-forward network with weights $\mW_1, \mW_2$, biases $w_1, w_2$, and a activation function $\sigma(\cdot)$, i.e.
$\text{FF}(\vx) = \sigma(\vx \mW_1 + w_1) \mW_2 + w_2.$
% \begin{equation}
%     \text{FF}(\vx) = \sigma(\vx \mW_1 + w_1) \mW_2 + w_2.
% \end{equation}
Aside from the residual selection, this is identical to common Transformer Encoder implementations~\citep{vaswani2017attention, dosovitskiy2021an}.

\subsection{Recycling Positional Encoding} \label{sec:circular-encoding}
Since a Transformer Encoder does not provide positional bias, it is common to augment a token $\vx_i$ with a positional encoding $\vp$, i.e. $\Tilde{\vx_i} = \vx_i \circ \vp_i$, where $\circ$ could be addition or concatenation.
% Though tokens often have inherent positional relations, their order in the (Continual) SDA does not in itself introduce the needed bias to model such relation. 
% Instead, it is common to augment a token, $\vx_i$, with a positional encoding $\vp$, i.e. $\Tilde{\vx_i} = \vx_i \circ \vp_i$, where $\circ$ could be addition or concatenation.
% which can be either predefined (e.g., sinusoidal) or learned~\citep{vaswani2017attention}.% (acting as an inter-token relation bias). 
% Prior to inputting a step vector $\vx_i$ into the Transformer Encoder Block, the vector would thus be augmented with the corresponding position $\vp_i$, i.e. $\Tilde{\vx_i} = \vx_i \circ \vp_i$,
% \begin{equation}
%     \Tilde{\vx_i} = \vx_i \circ \vp_i,
% \end{equation}
% where $\circ$ could be addition or concatenation.
%
% such encoding signifies a static position in the token sequence;
In regular Transformers, the index $i$ denotes a position in a sequence rather than a position in time.
% Such encoding signifies a static position in the token sequence. Specifically, the same positional encoding is used to augment the ``last'' input token of any input sequence.
However, this static positional assignment is problematic in the context of continual inference; the last token at time $t=0$ will be the next-to-last token at time $t=1$, and thus in need of a different positional encoding than in the prior time-step. Instead, CINs require dynamic positions.
% If we were to allow the positional encoding of prior tokens to change retroactively, the partial attention results of a subsequent Continual Scaled Dot-Product Attention would become invalid. 
%
There have been multiple prior works~\citep{Shaw2018SelfAttentionWR, Huang2019MusicTG, dai2019transfromerxl} which create relative encodings by augmenting the SDA with positional offsets $\mP$ between query and keys, i.e. $\mA = \text{exp}(\mQ \mK^{\top} / \sqrt{d} + \mP)$.
While a similar modification to the continual attentions is possible, it is incompatibile with the regular SDA in \cref{eq:scaled-dot-product-attention}.
Instead, we use a \textit{Recycling Positional Encoding} (RPE), which lets the positional encoding follow each token in time and recycles old encodings: % in a round-robin manner:
% Positional encodings corresponding to old tokens, which have slided out of the scope can be reused for subsequent tokens in a circular manner.
\begin{gather}
    \Tilde{\vx_t} = \vx_t + \vp_{\tau_t}, \label{eq:add-enc}
    \qquad
    \tau_t = (\tau_{t-1} + 1) \text{ mod } T,
\end{gather}
% We dub this approach \textit{Recycling Positional Encodings} (RPE).
%
where $T$ is the number of encodings.
While RPE does not specify relative encodings explicitly, the absolute positional interpretation of each token changes dynamically when a new token arrives. %; the positional encoding implicitly becomes relative.
In practice, the network learns relative, shift-invariant positional information by training with random $\tau_0$ for each batch. %RPE can be used for both batch inference and step-wise inference with the same results. 
Random shifts during training were recently explored in \citep{kiyono2021shape, likhomanenko2021cape, dehghani2018universal} as well.
RPE can use either learned or predefined encodings. In the latter case, Cyclic Positional Encoding~\citep{ma2021learning}, a sinusoidal 
encoding inspired by Gray code, is a good fit. 
If we reuse the encoding immediately after an old token has ``slided out'', i.e. $T=n$, a token will have the same positional encoding relative to another whether it was $m$ steps older or $n-m$ steps newer. 
% This strategy was also used by \cite{likhomanenko2021cape}. % used for the automatic speech recognition experiments in \citep{likhomanenko2021cape}.
The positional ambiguity can be avoided by extending the number of positional tokens to $T=2n-1$.
We explore both options in \cref{sec:exp-cooadtr}.
%

% We can either reuse a token that has ``slided out'' immediately in the next step or we can extend the number of positional tokens to $2n$ and wait $n$ steps before reusing a token again. 
% While the latter option ensures that non-ambiguous relative positions can be learned (in the former option, a token would have the same positional encoding relative to another whether it was $m$ steps older or $n-m$ steps newer), it comes with an increased parameter count.

\subsection{Architectural considerations} \label{sec:arch-considerations}
% A grounding principle of Continual Inference Networks is the restriction that step results should be equal to the partial results of batch-inference over the complete time-series. This is necessary to ensure that any learned weights are compatible in both regular and step-wise inference modes.
\paragraph{Block count}
In \cref{sec:co-comparison}, we observed an exact correspondence between the results of the continual and regular SDA layers. 
However, the correspondence does not necessarily hold for stacked layers. 
Consider the result of stacking two Continual Single-Output Transformer Encoder blocks.
While the first block outputs a step $t$ that is identical to that in a corresponding regular block, the second block would have been initialised with prior step-wise inputs, which were the result of prior input windows instead of the current one; the correspondence would not hold. 
Though it is not convertible to/from a regular Transformer Encoder, the stacked Single-Output Transformer Encoder architecture has the merit of efficiency. This was exploited in Transformer-XL~\citep{dai2019transfromerxl}. % to achieve competitive results at low step-wise inference cost. % It would, however, not be convertible to/from a regular Transformer architecture.
Given a single step input, the Continual Retroactive Transformer Encoder block produces output tokens corresponding to the entire observed sequence inside the window. Due to this one-to-many input-output mapping, it is not possible to stack multiple such layers. 
Nevertheless, it can be used in conjunction with a Continual Single-Output Transformer Encoder with optional regular Transformer Encoder blocks in between as illustrated in \cref{fig:three-layer-architecture}.
The Regular Transformer Encoder blocks in the resulting architecture have a significantly larger computational complexity than the Continual Retroactive and Single-Output blocks. 
Consequently, we recommend that Continual Transformer Encoders be used primarily in lightweight architectures with one or two blocks unless compatibility with non-continual Transformers is not required and only Single-output blocks are used.

\paragraph{Class token}
It is common to add a class token as input to transformers~\citep{devlin2019bert, dosovitskiy2021an}, which accumulates information from other tokens prior to classification.
However, it cannot be used naïvely with CINs, as this would effectively double the number of input steps. 
In practice, it can be employed in Continual multi-block Transformer Encoders as input to the second block (see \cref{fig:three-layer-architecture}), but this placement limits class token interaction with downstream layers. 
It can also be used for one-block Transformer Encoders if the value token is omitted as input.

\paragraph{Peak memory reduction trick}
% \subsubsection{Reducing the memory load of Regular Scaled Dot-Product Attention}
The FLOPs for $\text{Att}(\mQ, \mK, \mV)$ are exactly $n$ times those of $CoSi\text{Att}(\mathbf{q}, \vkn, \vvn)$. % and $Si\text{Att}(\mathbf{q}, \mK, \mV)$. 
Comparing their memory complexity, the regular SDA is $\mathcal{O}({n^2})$, while the Single-output SDA is $\mathcal{O}({nd})$. In practical applications where system memory is limited, we may thus reduce the maximum memory requirement of the computational device at inference by up to $d/n$ (assuming $n \gg d$) by computing each row of the attention individually. 
% Depending on the device at hand, 
However, this may reduce throughput due to reduced parallelism.

% \begin{figure}[t]
%     \centering
%     \includegraphics[width=\linewidth]{figures/TwoLayerCoReMHA.pdf}
%     \caption{
%         \textbf{Two-layer Continual Multi-Head Attention.} 
%     }
%     \label{fig:two-layer-architecture}
% \end{figure}

% \begin{figure}[t]
%     \centering
%     \includegraphics[width=\linewidth]{figures/SingleLayerCoReMHA.pdf}
%     \caption{
%         \textbf{Single-layer Continual (Retroactive) Multi-Head Attention with Circular Positional Encoding.} 
%     }
%     \label{fig:single-layer-architecture}
% \end{figure}

\section{Experiments}
% To investigate the applicability of Continual Transformers, w
We provide case studies within two perception disciplines, namely Online Action Detection (\cref{sec:exp-oad}) and Audio Classification (\cref{sec:exp-ac}). In each case, we will start with a brief overview of the field, followed by experiments and results.

% The continual modules proposed in \cref{sec:cotrans}, will be made available as stand-alone modules for PyTorch and TensorFlow, as well a unified repository for Continual Inference Networks\footnote{Library links will be added in final version.}.
% The continual modules proposed in \cref{sec:cotrans}, are available as stand-alone modules for PyTorch\footnote{\url{https://github.com/lukashedegaard/continual-transformers}} and TensorFlow\footnote{\url{https://github.com/lukashedegaard/continual-transformers-tf}}, and have been added to the unified repository for Continual Inference Networks\footnote{\url{https://github.com/lukashedegaard/continual-inference}}\citep{hedegaard2022colib}.

\subsection{Online Action Detection} \label{sec:exp-oad}
% \subsubsection{Background}
% Though the Transformer style neural network architecture was initially conceived for Natural Language Processing~\citep{vaswani2017attention}, it has since found widespread application in Computer Vision tasks such as Image Classification~\citep{dosovitskiy2021an}, Object Detection ~\citep{carion2020end}, Trimmed Human Action Recognition~\citep{Arnab2021ViViTAV}, and Online Action Detection~\citep{wang2021oadtr, xu2021long}.
Online Action Detection (OAD)~\citep{geest2016online} entails the per-frame classification of human actions in a video stream as they happen without the ability to change prior predictions nor to use future information. 
This is fundamentally more restrictive than Temporal Action Localisation, where the whole video clip is processed before start and end frames of an action are determined~\citep{shou2016temporal, xu2017rc3d, shou2017cdc, wu2019longterm}.

The dominant design in OAD works at the time of writing is to employ a two-stream Convolutional Neural Network as backbone for frame-wise feature extraction with RGB images as inputs in one stream and Optical Flow fields in the other~\citep{gao2017red, xu2019temporal, Eun2020LearningTD, wang2021oadtr, xu2021long}\footnote{
The feature extraction commonly used in Online Action Detection (OAD) works is in itself quite computationally costly. 
We consider the optimisation of the backbone as orthogonal future work and will follow the same feature extraction procedure as other OAD works at this time.}.
On top of these, OAD methods encode temporal information and perform predictions per time-step, e.g. by means of RNNs~\citep{gao2017red, xu2019temporal, Eun2020LearningTD} or Transformers~\citep{wang2021oadtr, xu2021long}. 
Alongside the action detection for the current frame, an action anticipation task may be learned in parallel by means of decoder structures, as this has been found to improve the primary OAD task.

Unlike RNNs, an output update for the regular SDA in a Transformer block cannot be naïvely computed for a single step by feeding successive video frames. Instead, prior step features must be cached, re-loaded and re-processed by the Transformer in each step in correspondence with a predefined window-size of prior steps. 
% \subsubsection{Setup} \label{sec:oad-setup}
As laid out in \cref{sec:arch-considerations}, Continual Transformers are especially efficient when either one or two Continual Transformer Encoder blocks are used. 
Accordingly, we start our experiments with a set ablation studies to simplify a recent transformer-based architecture, the OadTR~\citep{wang2021oadtr}. %, including an ablation of class token positions.
We further investigate the impact of ablating class token position and the use of Recycling Positional Encoding and compare different RPE schemes for Continual Transformers.
Finally, we evaluate our configurations on two widely used OAD datasets, THUMOS14~\citep{idrees2017thumos} and TVSeries~\citep{geest2016online}.%, and compare the attained performance with that of prior works.

\subsubsection{Experimental setup}\label{app:oad_setup}
The THUMOS14 dataset~\citep{idrees2017thumos} for OAD has 200 and 213 validation and test videos, respectively, with frame-level class annotations across 20 classes. 
As in prior OAD works, the model is trained on the validation set and evaluated on the test set. 
Similar to \cite{wang2021oadtr} we use pre-extracted features from a two-stream Temporal Segment Network (TSN)~\citep{wang2018tsn} trained on ActivityNet v1.3~\citep{heilbron2015activitynet} or Kinetics-400~\citep{carreira2017quo}.

For TVSeries~\citep{geest2016online}, the network learns on the train and validations sets (20 videos) and evaluates on the test set (7 videos) as in \citep{wang2021oadtr}.
RGB and Optical Flow features were extraced using an \mbox{MMAction2}~\citep{2020mmaction2} pipeline with ActivityNet v1.3~\citep{heilbron2015activitynet} and Kinetics-400~\citep{carreira2017quo} pretrained TSN ResNet-50~\citep{He2016} backbones. 
This is similar to the feature extraction process used by LSTR~\citep{xu2021long}.

Following \citet{wang2021oadtr}, we use a batch size of 128, sequence length 64, initial learning rate $10^{-4}$ with a factor ten reduction each epoch, alongside weight decay $10^{-4}$, and dropout with probability $0.1$. We report results using two epochs of training on a Nvidia RTX2080 Ti GPU.  % This is moved to apx:training-details
We track mean Average Precision (mAP) for THUMOS14 and calibrated mean Average Precision (cmAP)~\citep{geest2016online} for TVSeries, alongside FLOPs per prediction and parameters of the OAD module (feature extraction excluded). We report the mean $\pm$ standard deviation over five runs.
% Further training details are available in \cref{apx:training-details}.

\subsubsection{Ablation studies} \label{sec:exp-cooadtr}

\paragraph{Removing the Decoder}
As a first step to make an efficient Continual OadTR, we remove the decoder blocks used for action anticipation, which has a large impact on computational efficiency and the ease of transformation to a Continual Inference Network.
The first two lines of \cref{tab:oad-ablation} present the results of the removal.
Contrary to the observations of \citet{wang2021oadtr}, we did not find any drop in accuracy when excluding the decoder.
We do, however, gain a large reduction in FLOPs and model size; they were reduced to 58\% and 30\%, respectively.
Given these computational improvements, we exclude the decoder in subsequent experiments.

\paragraph{(Re)moving the Class token}
Class tokens should not be input naively to the first Transformer Encoder layer of a CIN (see \cref{sec:arch-considerations}).
Accordingly, we ablate its use and position. %experiment with introducing class token later or removing it altogether. 
In cases where it is removed, we predict on the token corresponding to the last input token. 
The results of varying \texttt{CLS} pos are noted in \cref{tab:oad-ablation}. % and visualised in \cref{fig:oad-cls-ablation}.
For the one-block architecture, the removal came with noticeable drop in mAP, while the two-block architecture saw small improvements when removing or introducing the class token later.
For the three block model, the use of class tokens in block two achieved the highest mAP. 
Though it is commonly accepted, that class tokens should be introduced alongside other inputs in the first block, our results indicate that they can accumulate sufficient information with only one or two blocks, and that later stage introduction may work better in some applications.
In general, the achieved mAP when varying \texttt{CLS} pos. and number of blocks are very similar to one another, while (re)moving the class token and reducing the block size both reduce computational complexity.
This encourages the use of shallow Transformer Encoders over deeper ones as well as the removal of class tokens, as we do in the following experiments.

\begin{table}
\caption{
    \textbf{Ablation experiments} on THUMOS14 with TSN-Anet features. 
    \best{Best} metrics are highlighted.
    \mbox{`-'} indicates that a particular feature was not used.
}
\begin{subtable}[h]{0.44\textwidth}
\caption{
    \textbf{Class token} variations with OadTR. 
    \texttt{CLS} pos. is the encoder block into which \texttt{CLS} is input. 
}\label{tab:oad-ablation}
\resizebox{\linewidth}{!}{
\begin{tabular}{cccccc}
    \toprule

    \textbf{Enc.}   & \multirow{2}{*}{\textbf{Dec.}} & \textbf{\texttt{CLS}}   & \textbf{mAP}    & \textbf{FLOPs}  & \textbf{Params}   \\
    \textbf{blocks}     &                                   & \textbf{pos.}    & \textbf{(\%)}   & \textbf{(M)}    & \textbf{(M)}  \\

    \midrule
    3                   & \checkmark    & 1         & 57.0\tiny{$\pm$0.5} & 2445.6          & 74.7  \\
    3                   & -             & 1         & 57.0\tiny{$\pm$0.4} & 1430.6          & 22.2  \\
    % 3                   & -             & 1F        & 57.0\tiny{$\pm$0.4} & 1425.6          & 22.2  \\
    3                   & -             & 2         & \best{57.3\tiny{$\pm$0.7}} & 1423.5          & 22.2  \\
    % 3                   & -             & 2F        & 57.0\tiny{$\pm$0.6} & 1419.3          & 22.2  \\
    3                   & -             & 3         & 56.7\tiny{$\pm$0.6} & 1417.2          & 22.2  \\
    3                   & -             & -         & 56.8\tiny{$\pm$0.3} & \best{1410.9}           & 22.2  \\
    \midrule
    2                   & -             & 1         & 56.5\tiny{$\pm$0.5} & 1020.7          & 15.9  \\
    % 2                   & -             & 1F        & 56.2\tiny{$\pm$0.3} & 1016.6          & 15.9  \\
    2                   & -             & 2         & \best{56.7\tiny{$\pm$0.3}} & 1014.5          & 15.9  \\
    2                   & -             & -         & 56.6\tiny{$\pm$0.3} & \best{1008.1}          & 15.9  \\
    \midrule
    1                   & -             & 1         & \best{57.1\tiny{$\pm$0.6}} & 611.7           & 9.6   \\
    1                   & -             & -         & 56.3\tiny{$\pm$0.2} & \best{605.5}           & 9.6   \\

    \bottomrule
\end{tabular}
}
\end{subtable}
\hfill
\begin{subtable}[h]{0.53\textwidth}
\caption{
    \textbf{Positional encodings} variations for \textit{Co}OadTr.
}\label{tab:cooadtr-configs}
\resizebox{\linewidth}{!}{
\begin{tabular}{ccccccc}
    \toprule

    \textbf{Enc.}  & \textbf{Re-}    & \multirow{2}{*}{\textbf{Learn}} & \textbf{Pos.} & \textbf{mAP}  & \textbf{FLOPs}     & \textbf{Params}   \\
    \textbf{blocks}    & \textbf{cycling}& & \textbf{tokens} & \textbf{(\%)} & \textbf{(M)}      & \textbf{(K)}  \\
    \midrule
    2   & -             & \checkmark & $n$           & 45.3\tiny{$\pm$0.9}  & 410.9  & 15832  \\
    2   & \checkmark    & \checkmark & $n$           & 56.4\tiny{$\pm$0.3}  & 410.9  & 15832  \\ 
    2   & \checkmark    & \checkmark & $2n$$-$$1$    & 56.0\tiny{$\pm$0.5}  & 410.9  & 15897  \\ 
    2   & \checkmark    & -          & $n$           & 55.8\tiny{$\pm$1.0}  & 410.9  & \best{15767}   \\ 
    2   & \checkmark    & -          & $2n$$-$$1$    & \best{56.8\tiny{$\pm$0.4}}  & 410.9  & \best{15767}   \\ 
    \midrule
    1   & -             & \checkmark & $n$           & 44.0\tiny{$\pm$0.8}  & 9.6   & 9535  \\
    1   & \checkmark    & \checkmark & $n$           & 55.6\tiny{$\pm$0.3}  & 9.6   & 9535  \\ 
    1   & \checkmark    & \checkmark & $2n$$-$$1$    & 55.6\tiny{$\pm$0.3}  & 9.6   & 9599  \\ 
    1   & \checkmark    & -          & $n$           & 54.4\tiny{$\pm$1.8}  & 9.6   & \best{9469}  \\ 
    1   & \checkmark    & -          & $2n$$-$$1$    & \best{56.1\tiny{$\pm$0.7}}  & 9.6   & \best{9469}  \\ 
    \bottomrule
\end{tabular}
}
\end{subtable}
\end{table}

\paragraph{Positional Encodings}
We can transfer parameters from the simplified one- and two block OadTR to the corresponding Continual architecture, \textit{Co}OadTR.
Here, the one block version (\textit{Co}OadTR-b1) uses \textit{CoSi}MHA, and the two block model (\textit{Co}OadTR-b2) uses \textit{CoRe}MHA in the first block and Single-output MHA in the second.
However, a regular positional encoding is not suited for continual inference (see \cref{sec:circular-encoding}).  
We evaluate the performance of using non-continual encodings for continual inference, as well as of our proposed Recycling Positional Encodings with fixed or learned parameters. 
In addition, we explore the impact of extending the number of tokens from $n$ to $2n-1$ to avoid positional ambiguity.
As seen in \cref{tab:cooadtr-configs}), non-continual encoding used in the continual setting result in severe mAP drop.
Recycling Positional Encodings alleviate this. 
Comparing learned and fixed encodings, we find the learned encodings to work better when the number of encoding tokens corresponds to the sequence length $n$ and the fixed encoding to work best when positional ambiguity is alleviated by extending the number of tokens to $2n-1$. Fixed encoding with $2n-1$ tokens works best overall and is employed in subsequent experiments unless stated otherwise. There is no difference in FLOPs for either strategy, and the difference in parameter count is negligible.

\subsubsection{Comparison with prior works}
\begin{figure}%{\textwidth}
\noindent\begin{minipage}{.59\linewidth}
	\begin{center}
	\captionof{table}{
	    \textbf{Online Action Detection} results.
	   % \textbf{Comparison} with prior Online Action Detection works. 
	    FLOPs per prediction are noted for inference on THUMOS14. 
	    The \best{best} and \nextbest{next-best} metrics are highlighted.
	}\label{tab:oad-prior-works}
    \resizebox{\linewidth}{!}{
	\begin{tabular}{llcccc}
		\toprule
		\multirow{2}{*}{\textbf{Model}} & \multirow{2}{*}{\textbf{Feat.}}  & \textbf{THUMOS14} & \textbf{TVSeries}     & \textbf{FLOPs} \\
		                                &                                  & \textbf{mAP (\%)}          & \textbf{mcAP (\%)}              &\textbf{(M)} \\
		\midrule
		RED~\citep{gao2017red}           &\multirow{12}{*}{A.Net}& 45.3\phantom{\tiny{$\pm$0.0}}             & 79.2\phantom{\tiny{$\pm$0.0}}                & - \\
		TRN~\citep{xu2019temporal}       &                       & 47.2\phantom{\tiny{$\pm$0.0}}             & 83.7\phantom{\tiny{$\pm$0.0}}                & 1387.5 \\ 
		    % flops: 88,800,113,024 for all steps, 1,387,501,766 for a single; params: 357,828,630
		FATS~\citep{kim2021temporally}   &                       & 51.6\phantom{\tiny{$\pm$0.0}}             & 81.7\phantom{\tiny{$\pm$0.0}}                & - \\
		IDN~\citep{Eun2020LearningTD}    &                       & 50.0\phantom{\tiny{$\pm$0.0}}             & 84.7\phantom{\tiny{$\pm$0.0}}                & - \\
        TFN~\citep{eun2021temporal}      &                       & 55.7\phantom{\tiny{$\pm$0.0}}             & 85.0\phantom{\tiny{$\pm$0.0}}                & - \\
% 		LFB~\citep{wu2019long,xu2021long}&                       & 61.6\phantom{\tiny{$\pm$0.0}}             & 84.8\phantom{\tiny{$\pm$0.0}}                & - \\
		LSTR~\citep{xu2021long}          &                       & \best{65.3}\phantom{\tiny{$\pm$0.0}}             & 88.1\phantom{\tiny{$\pm$0.0}}                & - \\
		OadTR~\citep{wang2021oadtr}      &                       & \nextbest{58.3}\phantom{\tiny{$\pm$0.0}}             & 85.4\phantom{\tiny{$\pm$0.0}}                & 2445.6 \\
% 		\midrule
        OadTR$^\dagger$                 &                       & 57.0\tiny{$\pm$0.5}             & \best{88.6}\tiny{$\pm$0.1}               & 2445.6 \\
	    OadTR-b2$^\dagger$              &                       & 56.6\tiny{$\pm$0.3}             & \nextbest{88.3\tiny{$\pm$0.2} }                    & 1008.1 \\
	    OadTR-b1$^\dagger$              &                       & 56.3\tiny{$\pm$0.2}             & 88.1\tiny{$\pm$0.1}                     & 605.5 \\
		\textit{Co}OadTR-b2 (ours)      &                       & 56.8\tiny{$\pm$0.4}             & 87.7\tiny{$\pm$0.6}                     & \nextbest{410.9} \\%($-2.5\times$) \\
		\textit{Co}OadTR-b1 (ours)      &                       & 56.1\tiny{$\pm$0.7}             & 87.6\tiny{$\pm$0.7}                    & \best{9.6} \\% ($-63.1\times$) \\

		\midrule

        TRN~\citep{xu2019temporal}       &                       & 62.1\phantom{\tiny{$\pm$0.0}}             & 86.2\phantom{\tiny{$\pm$0.0}}                & 1462.0 \\
            % flops: 93,566,022,016 for all steps, 1,461,969,094 for a single; params: 402,919,446
		FATS~\citep{kim2021temporally}   & \multirow{10}{*}{Kin.}& 59.0\phantom{\tiny{$\pm$0.0}}             & 84.6\phantom{\tiny{$\pm$0.0}}                & - \\
		IDN~\citep{Eun2020LearningTD}    &                       & 60.3\phantom{\tiny{$\pm$0.0}}             & 86.1\phantom{\tiny{$\pm$0.0}}                & - \\
% 		LFB~\citep{wu2019long,xu2021long}&                       & 64.8\phantom{\tiny{$\pm$0.0}}             & 85.8\phantom{\tiny{$\pm$0.0}}                & - \\
		PKD~\citep{Zhao2020PrivilegedKD} &                       & 64.5\phantom{\tiny{$\pm$0.0}}             & 86.4\phantom{\tiny{$\pm$0.0}}                & - \\
		{LSTR}~\citep{xu2021long}        &                       & \best{69.5}\phantom{\tiny{$\pm$0.0}}             & \best{89.1}\phantom{\tiny{$\pm$0.0}}                & - \\
		OadTR~\citep{wang2021oadtr}      &                       & \nextbest{65.2}\phantom{\tiny{$\pm$0.0}}             & 87.2\phantom{\tiny{$\pm$0.0}}                & 2513.5 \\
        % \midrule
        {OadTR$^\dagger$}               &                       & 64.2\tiny{$\pm$0.3}             & \nextbest{88.6\tiny{$\pm$0.1}}              & {2513.5} \\ % 2,513,543,748
	    OadTR-b2$^\dagger$              &                       & 64.5\tiny{$\pm$0.5}             & 88.3\tiny{$\pm$0.2}               & 1075.7 \\ %  1,075,714,070
	    OadTR-b1$^\dagger$              &                       & 63.9\tiny{$\pm$0.5}             & 88.1\tiny{$\pm$0.1}               & 673.0 \\ % 672,796,182
		{\textit{Co}OadTR-b2 (ours)}    &                       & 64.4\tiny{$\pm$0.1}             & 87.6\tiny{$\pm$0.7}               & \nextbest{411.9} \\ % 411,854,872, 64.1 for CPE 2n
		{\textit{Co}OadTR-b1 (ours)}    &                       & 64.2\tiny{$\pm$0.4}             & 87.7\tiny{$\pm$0.4}               & \best{10.6 }\\ % 
		\bottomrule
	\end{tabular}
	}
	\end{center}
	{\footnotesize$^\dagger$Using official source code or modifications there-off.
% 	The b1/b2 suffix refers to a simplified one/two block architecture without decoder and class token.
	}

\end{minipage}%
\hfill
\begin{minipage}{.38\linewidth}

    \begin{center}
    
    \includegraphics[width=\linewidth]{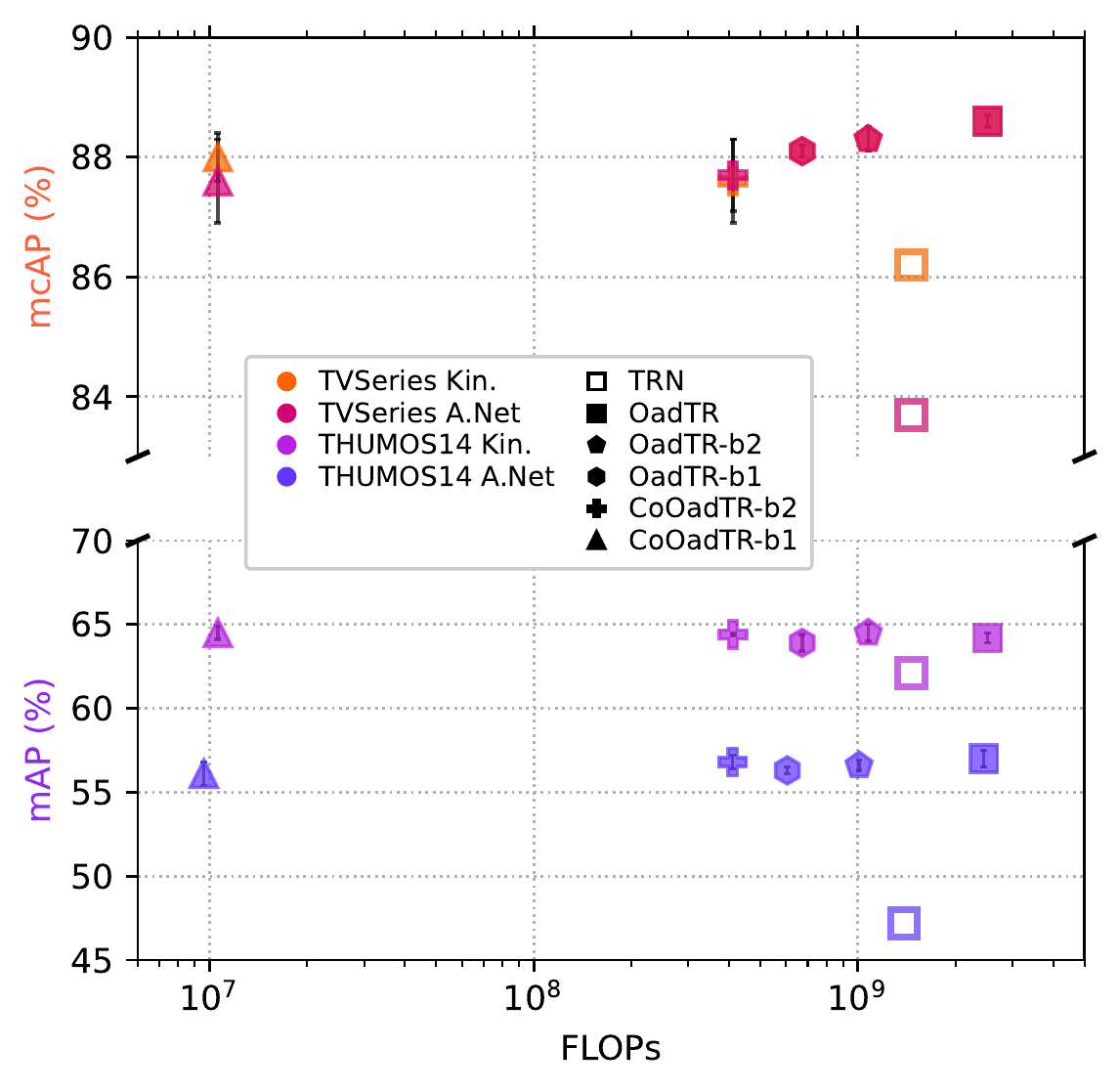}
    \captionof{figure}{
        \textbf{Visual comparison} of OAD methods on \mbox{THUMOS14} and \mbox{TVSeries} for backbones trained on \mbox{ActivityNet 1.3} and \mbox{Kinetics-400}.
    }\label{fig:oad-comparison}
	
	\end{center}

\end{minipage}

\end{figure}

We evaluate the (\textit{Co})OadTR architectures on THUMOS14 and TVSeries with two sets of features as described in \cref{app:oad_setup}. 
% In addition to tracking precision metrics, we measure the Floating Point Operations (FLOPs) per prediction for each of our model variants. 
Since no prior OAD works have reported complexity metrics, we measured the FLOPs for TRN~\citep{xu2019temporal} based on the publicly available source code to serve as a point of reference. 
The results of this benchmark are presented in \cref{tab:oad-prior-works} and \cref{fig:oad-comparison}.
OadTR and our simplified (continual) one-block (b1) and two-block (b2) versions without decoder and class tokens generally achieve competitive precision in comparison with prior works, surpassing all but OadTR and LSTR. 
On \mbox{THUMOS14}, our reproduced OadTR results are slightly lower than originally reported~\citep{wang2021oadtr}\footnote{
% While a single run did reached 57.8\% mAP (another as low as 56.4\%), 
The reported 58.3\% on THUMOS14 could not be reproduced using their publicly available code.
}, whereas achieved TVSeries results are higher\footnote{We attribute our higher mcAP to differences in the feature extraction pipeline.}.
The (\textit{Co})OadTR-b\# architecture largely retain precision and allow significantly reduced FLOPs per prediction. 
Our proposed continual variants \textit{Co}OadTR-b1 and \textit{Co}OadTR-b2 reduce FLOPs by \textbf{255$\times$} and \textbf{6.1$\times$}, respectively, compared to OadTR, while either achieving the same performance or conceding no more than one percentage point.
On average, continual and non-continual (\textit{Co})OadTR-b\# models achieve similar mAP on THUMOS14, while OadTR-b\# have slightly higher mcAP on TVSeries. We attribute these discrepancies to differences in positional encoding.
All in all, the \mbox{\textit{Co}OadTR-b\#} models provide far-superior computational efficiency to prior works, achieving state-of-the-art performance/efficiency trade-offs by a large margin.

\subsubsection{Audio-Visual Online Action Detection}
To showcase the validity of our method in audio-visual settings as well, we explore the addition of audio-features to the Online Action Detection task on THUMOS14.
As described in \cref{sec:exp-ac}, audio-features are extracted using Mel spectrograms and an AudioSet pre-trained VGGish network~\citep{Hershey2017} (output of the penultimate layer) on 1.0 second windows with a step size of 0.2 seconds to match the 5.0 FPS sampling rate of the video features. 
% The feature extraction pipeline is described in more details in \cref{sec:exp-ac}.
% In order to extract the audio features, initially, raw audio waveforms are extracted from each video. Then, smaller waveforms are extracted by sliding a window of size 1s with step size of 0.2s on the larger waveform. Subsequently, the 1-second waveforms are converted to Mel spectrograms, which are explained in greater detail in section~\ref{sec:exp-ac}. Finally, the spectrograms are given as input to a VGGish network pre-trained on AudioSet~\citep{Hershey2017} and the output of the penultimate layer of the network is used as audio features.

The audio-features by themselves do not provide enough signal to reach good Online Action Detection performance (yielding only $6.7\%$ mAP with an OadTR network).
When concatenated with RGB and Flow they do provide a modest improvement as seen in \cref{tab:oad-audio-visual}.
On average, this amounts to $+0.6\%$ mAP when combined with ActivityNet features and $+0.5\%$ mAP when used with Kinetics-400 features with shallower models enjoying the largest improvements. %These slight performance increases do however come at an increased computational cost.

\subsection{Audio Classification} \label{sec:exp-ac}

\subsubsection{Background}
Audio Classification is the categorisation of audio waveforms. Though waveform sequences can be used directly~\citep{1712.00866}, it is common to first convert them to spectrograms. Mel spectrograms are obtained by a nonlinear transformation of a frequency scale \citep{doi:10.1121/1.1915893}, which is designed based on empirical knowledge about the human auditory system~\citep{1606.00298}. By employing spectrograms, audio classification can be approached in the same way as image classification~\citep{2007.11154}.

\subsubsection{Experiments}
We conduct experiments on the Music Genre Classification dataset GTZAN~\citep{Tzanetakis2002}. 
It consists of 100 30-second clips for each of ten music genres. 
Each audio clip is sampled at 22,050 Hz. 
Since there are no predefined splits for GTZAN, we randomly select 10\% of the data for validation and 10\% for testing. 
The input is transformed to a temporal sequence by sliding a one-second window over each 30-second clip with a slide step size of 250ms, leading to 120 one-second clips. % where each clip has a 75\% overlap with the previous and next clips. 
These are subsequently converted to Mel spectrograms. 
We then fine-tune a VGGish network, pre-trained on AudioSet~\citep{Hershey2017} and use the penultimate layer for feature extraction.
% Further training details can be found in \cref{apx:training-details}.
A batch size of 64 and the Adam optimizer~\citep{DBLP:journals/corr/KingmaB14} are used with an initial learning rate of $10^{-4}$. The learning rate is reduced by a factor of $0.6$ on plateau with a tolerance of two epochs, and an early stopping mechanism, % with a tolerance of five epochs is employed, 
where a maximum of 100 epochs are allowed. 
The VGGIsh base-network attains an accuracy of 86.1\% on the dataset of one-second clips with 72.1M parameters and 864.7M FLOPs.
Subsequently, the audio features are passed to a (Continual) Transformer Encoder which has 16 attention heads, an embedding dimension of 192 and an MLP dimension of 384. The Transformer Encoder is trained on the whole temporal sequence using a batch size of 32 and the AdamW optimizer \citep{loshchilov2018decoupled} with a learning rate of $ 10^{-5} $ and a weight decay of $ 10^{-4} $ for 50 epochs. Since the Transformer Encoder is trained on entire 30-second clips, there are less data points available for this training. Accordingly, the size of the validation set is increased to 18\%. All audio classification training procedures were carried out on a single Nvidia RTX 2080 Ti GPU.
\cref{tab:audio_cls} presents the accuracy and efficiency of regular and Continual Transformers during online inference. As a baseline, we also include the result of majority voting among the clips to classify the entire sequence. The Continual Transformers obtain similar accuracy as regular a Transformers while consuming $ 1.76\times $ less FLOPs when using two blocks and $51.5\times $ less FLOPs when using one Transformer Encoder block.

\begin{figure}%{\textwidth}
\noindent\begin{minipage}{.45\linewidth}
    % \vspace{2em}
	\begin{center}
	\captionof{table}{
	    \textbf{Audio-Visual} result, \mbox{THUMOS14}.
	   % The \best{best} and \nextbest{next-best} metrics are highlighted.
	}\label{tab:oad-audio-visual}
    \resizebox{\linewidth}{!}{
	\begin{tabular}{lcccc}
		\toprule
		\textbf{Model}  & \textbf{Feat.}    & \textbf{mAP (\%)} & \textbf{FLOPs (M)}    \\
		\midrule
		OadTR           &                   & \best{57.6\tiny{$\pm$0.6}}       & 2714.9 \\ % +0.6
	    OadTR-b2        & A.Net             & \nextbest{57.5\tiny{$\pm$0.5}}   & 1277.0 \\ % +0.9
	    OadTR-b1        & +                 & 57.4\tiny{$\pm$0.4}              & 874.1 \\  % +1.3
		\textit{Co}OadTR-b2 & AudioSet      & 56.5\tiny{$\pm$1.1}               & \nextbest{415.0} \\ % -0.3
		\textit{Co}OadTR-b1 &               & 56.8\tiny{$\pm$0.5}              & \best{13.8} \\ % +0.7
		
		\midrule
		
  		OadTR           &                   & 64.4\tiny{$\pm$0.4}               & 2781.9 \\ % +0.2
	    OadTR-b2        & Kin.              & \best{65.0\tiny{$\pm$0.4}}        & 1344.1 \\ % +0.6
	    OadTR-b1        & +                 & 64.5\tiny{$\pm$0.4}               & 941.2 \\ % +0.6
		\textit{Co}OadTR-b2 & AudioSet      & 64.7\tiny{$\pm$0.8}               & \nextbest{416.0} \\ % +0.3
		\textit{Co}OadTR-b1 &               & \nextbest{64.8\tiny{$\pm$0.3}}   & \best{14.8} \\ % +0.6
		\bottomrule
	\end{tabular}
	}
	\end{center}
\end{minipage}
\hfill
\noindent\begin{minipage}{.45\linewidth}
	\begin{center}

    \captionof{table}{\textbf{Audio Classification} results for GTZAN.}\label{tab:audio_cls}
    \resizebox{\linewidth}{!}{
    \begin{tabular}{ lcccc }
        \toprule
        
        % \textbf{Method} & \textbf{Pos. Enc.} & \textbf{Acc. (\%)} & \textbf{FLOPs (M)} & \textbf{Par. (K)} \\
        \multirow{2}{*}{\textbf{Method}} & \multirow{2}{*}{\textbf{Pos. Enc.}}  & \textbf{Acc.} & \textbf{FLOPs} & \textbf{Par.} \\
                                         &                                      & \textbf{(\%)} & \textbf{(M)}      & \textbf{(K)} \\
        \midrule
    
        Maj. Voting & -  & 92.0\phantom{\tiny{$\pm$0.0}} & - & \best{0} \\
        % Regular 2-Layer Trans. & 94.0 & 47.4 & 508.6 \\
        % Regular 1-Layer Trans. & 94.0 & 15.2 & \textbf{286.3} \\
        % \textit{CoRe} + \textit{CoSi} Trans. & \textbf{95.0} & 27.0 & 508.6\\
        % \textit{CoSi} Trans. & 94.0 & \textbf{0.3} & \textbf{286.3} \\
        Trans-b2 & learned & \best{95.0\tiny{$\pm$0.6}} & 47.4 & 509\\
        Trans-b1 & learned & 93.8\tiny{$\pm$0.8} & 15.2 & \nextbest{286}\\
        \textit{Co}Trans-b2 & fixed & \nextbest{94.4\tiny{$\pm$1.0}} & 27.0 & 509\\
        \textit{Co}Trans-b1 & learned & 93.2\tiny{$\pm$1.1} & \nextbest{0.3} & \nextbest{286}\\
    
        \bottomrule
    \end{tabular}
    }
	\end{center}
\end{minipage}
\end{figure}

% \begin{table}[!htbp]
% \caption{\textbf{GTZAN} accuracy, FLOPs and parameter count using majority voting, regular Transformer Encoder blocks and Continual Transformer Encoder blocks.}\label{tab:audio_cls}
% \begin{center}
% \begin{tabular}{ l c c c c }
%     \toprule

%     \textbf{Method} & \textbf{Learned Pos. Enc.} & \textbf{Acc. (\%)} & \textbf{FLOPs (M)} & \textbf{Params (K)} \\

%     \midrule

%     Majority Voting & -  & 92.0 & - & - \\
%     % Regular 2-Layer Trans. & 94.0 & 47.4 & 508.6 \\
%     % Regular 1-Layer Trans. & 94.0 & 15.2 & \textbf{286.3} \\
%     % \textit{CoRe} + \textit{CoSi} Trans. & \textbf{95.0} & 27.0 & 508.6\\
%     % \textit{CoSi} Trans. & 94.0 & \textbf{0.3} & \textbf{286.3} \\
%     Trans-b2 & \checkmark & \textbf{95.0\tiny{$\pm$0.6}} & 47.4 & 508.6\\
%     Trans-b1 & \checkmark & 93.8\tiny{$\pm$0.8} & 15.2 & \textbf{286.3}\\
%     \textit{Co}Trans-b2 & - & 94.4\tiny{$\pm$1.0} & 27.0 & 508.6\\
%     \textit{Co}Trans-b1 & \checkmark & 93.2\tiny{$\pm$1.1} & \textbf{0.3} & \textbf{286.3}\\

%     \bottomrule
% \end{tabular}
% \end{center}
% \end{table}

% \subsection{Facial Expression Recognition}
% Training time, hardware, epochs, layers, hparams, dataset(s)

\section{Conclusion} \label{sec:conclusion}
In this work, we presented Continual Transformers, a redundancy-free reformulation of Transformers tailored for online inference. 
Central to the Continual Transformer are the Continual Retroactive and Single-Output Attention operations, which produce outputs identical to the original Scaled Dot-Product Attention for continual input sequences, while greatly reducing the time and memory complexity per prediction. 
The applicability of Continual Transformer architectures was experimentally validated in Online Action Detection and Online Audio Classification settings, observing upwards of multiple orders of magnitude reduction in time complexity for lightweight architectures at modest accuracy concessions.
Continual Transformers constitute an algorithmic innovation, which could make possible hitherto unseen precision, speed, and power efficiency in online inference use-cases. With applications spanning enhanced perception and reactivity of robots and autonomous vehicles, weather forecasting, price prediction and surveillance, %, this technology entails both the risks and upsides associated with the above.
we hope it will be used for the common good. 

% This work adds Transformers to the literature on Continual Inference Networks, which we hope will make viable new frontiers of latency-critical and recourse-constrained applications.


\subsubsection*{Acknowledgments}
This work has received funding from the European Union’s Horizon 2020 research and innovation programme under grant agreement No 871449 (OpenDR).

\bibliography{iclr2023_conference}
\bibliographystyle{iclr2023_conference}

\appendix
\section{Appendix}
% \subsection{Further description of the problem setting}
% \begin{figure}[t]
%     \centering
%     \includegraphics[width=0.4\linewidth]{figures/RedundancyIllustration-alt-colors.pdf}
%     \caption{
%         \textbf{Redundant computations during online inference} for a regular two-layer neural network with temporal connectivity and receptive field of three.
%         Contributing connections for the \textcolor{amber}{prior step $t_{-1}$} and \textcolor{ioite}{current step $t_{0}$} alongside \mbox{\textcolor{pink}{redundant}} computations are highlighted.
%     }
%     \label{fig:redundancy}
% \end{figure}

\subsection{Scaling properties of Continual and regular Multi-head Attention}\label{apx:mha-scaling}

A detailed account for the floating point operations involved in computing Regular-, Continual Retroactive-, and Single-output Scaled Product Attentions is given in Tables \ref{tab:dot-prod-complexity}, \ref{tab:core-dot-prod-complexity}, and \ref{tab:resi-dot-prod-complexity}.

\begin{table}[!htbp]
\caption{
    \textbf{Floating Point Operations} for the Scaled Dot-Product Attention in
    \cref{eq:scaled-dot-product-attention}. $\mD^{-1} (\cdot)$ can be efficiently computed as element-wise multiplication with $\mA \mV$.
}\label{tab:dot-prod-complexity}
\begin{center}
% \resizebox{\linewidth}{!}{
\begin{tabular}{l|ccc}
    \toprule
        & Mul.              & Add               & Exp       \\
    \midrule
    Eq. (\ref{eq:scaled-dot-product-attention}.1) 
        & $n^2d + nd$       & $nd(n-1)$      & 0 \\
    % & $\mA \mV$           & $n^2d$        & $n^2(d-1)$        &           \\
    % & $\mD^{-1} (\cdot)$  & $n+nd$          &                   &           \\
    Eq. (\ref{eq:scaled-dot-product-attention}.2) 
        & $n^2d + nd$       & $n^2(d-1)$    & $n^2$ \\
    % & $\mK / \sqrt{d}$    & $nd$          &                   &           \\
    % & $\mQ \mK^{\top}$    & $n^2d$        & $n^2(d-1)$        &           \\
    % & $exp(\cdot)$        &               &                   & $n^2$     \\
    Eq. (\ref{eq:scaled-dot-product-attention}.3) 
        & $0$               & $n(n-1)$      &  0 \\
    \bottomrule
\end{tabular}
% }
\end{center}
\end{table}

\begin{table}[!htbp]
\caption{
    \textbf{Floating Point Operations} for the Continual Retroactive Dot-Product Attention in
    \crefrange{eq:core-sdpa-d-update}{eq:core-sdpa-dav}.
    The outputs of the exponentials in \cref{eq:core-sdpa-d-update} and \cref{eq:core-sdpa-d-0} can be reused in \cref{eq:core-sdpa-av-update} and \cref{eq:core-sdpa-av-0} respectively, and are omitted in the count.
}\label{tab:core-dot-prod-complexity}
\begin{center}
% \resizebox{\linewidth}{!}{
\begin{tabular}{l|ccc}
    \toprule
    &           Mul.            & Add                   & Exp       \\
    \midrule
    \cref{eq:core-sdpa-d-update} 
                & $2(n-1)d$     & $2(n-2)d + 2(n-1)$    & $2(n-1)$  \\
    \cref{eq:core-sdpa-d-0} 
                & $nd + n + d$  & $nd + (n-1) + d$      & $n$       \\
    \cref{eq:core-sdpa-av-update} 
                & $2(n-1)d$     & $2(n-1)d$             & 0         \\
    \cref{eq:core-sdpa-av-0} 
                & $nd$          & $(n-1)d$              & 0         \\
    \cref{eq:core-sdpa-dav} 
                & $nd+n$        & 0                     & 0         \\

    \bottomrule
\end{tabular}
% }
\end{center}
\end{table}

\begin{table}[!htbp]
\caption{
    \textbf{Floating Point Operations} for the Continual Single-Output SDA in \cref{eq:leading-scaled-dot-product-attention-att}.
}\label{tab:resi-dot-prod-complexity}
\begin{center}
% \resizebox{\linewidth}{!}{
\begin{tabular}{l|ccc}
    \toprule
    &           Mul.            & Add                   & Exp       \\
    \midrule
    Eq. (\ref{eq:leading-scaled-dot-product-attention-att}.1) 
                & $nd + d$     & $(n-1)d + n - 1$       & $0$  \\
    Eq. (\ref{eq:leading-scaled-dot-product-attention-att}.2) 
                & $nd + d$  &   $n(d-1)$                & $n$       \\
    \bottomrule
\end{tabular}
% }
\end{center}
\end{table}

\begin{figure}
    \centering
	\begin{subfigure}[b]{0.325\linewidth}
        \includegraphics[width=\linewidth]{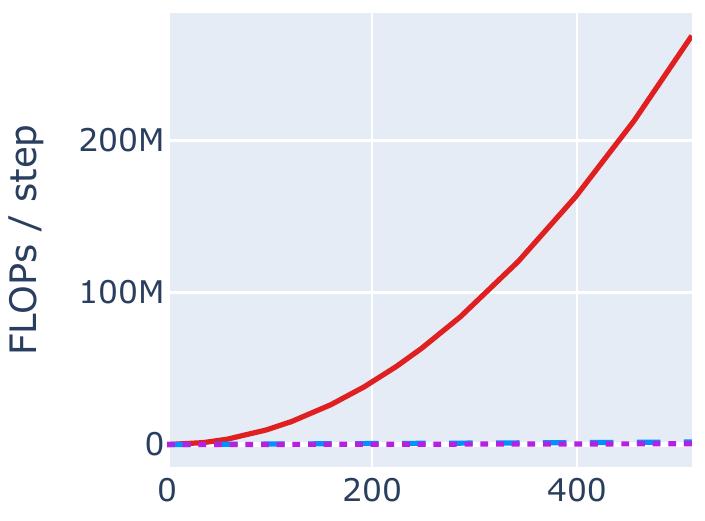}
        % \caption{$d = 100$}
        \label{fig:theory-attn-comp-fixed-e}
	\end{subfigure}
	\qquad
	\begin{subfigure}[b]{0.30\linewidth}
	    \includegraphics[width=\linewidth]{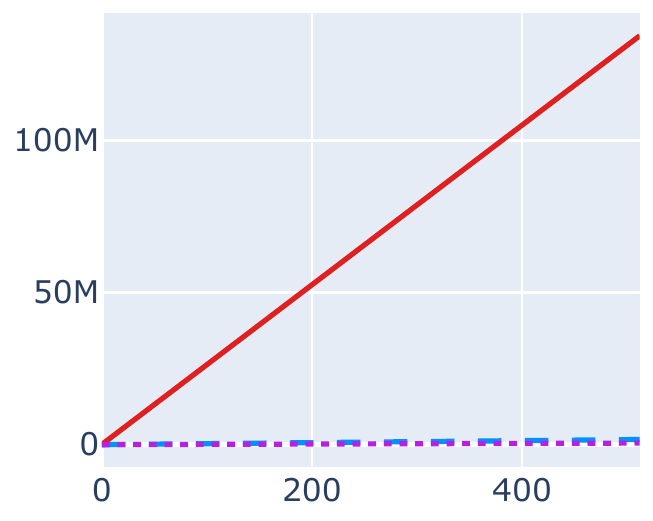}
	   % \caption{$n = 100$}
		\label{fig:theory-attn-comp-fixed-n}
	\end{subfigure}
	\\
	\begin{subfigure}[b]{0.325\linewidth}
        \includegraphics[width=\linewidth]{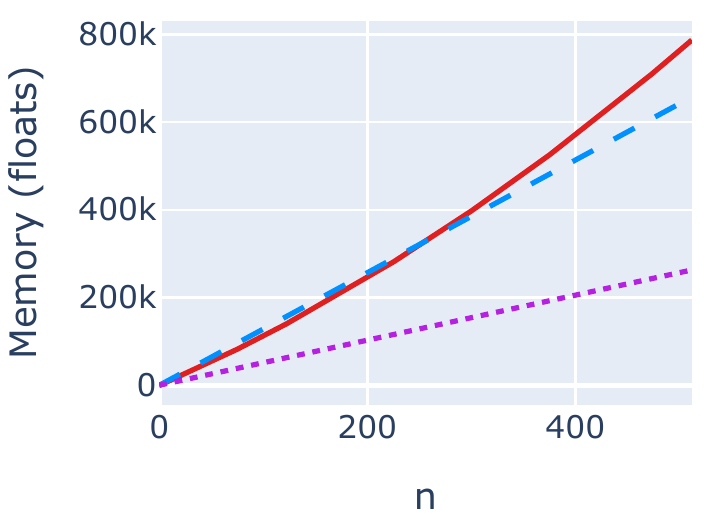}
        \caption{$d = 256$}
        \label{fig:theory-attn-comp-fixed-e-mem}
	\end{subfigure}
	\qquad
	\begin{subfigure}[b]{0.30\linewidth}
	    \includegraphics[width=\linewidth]{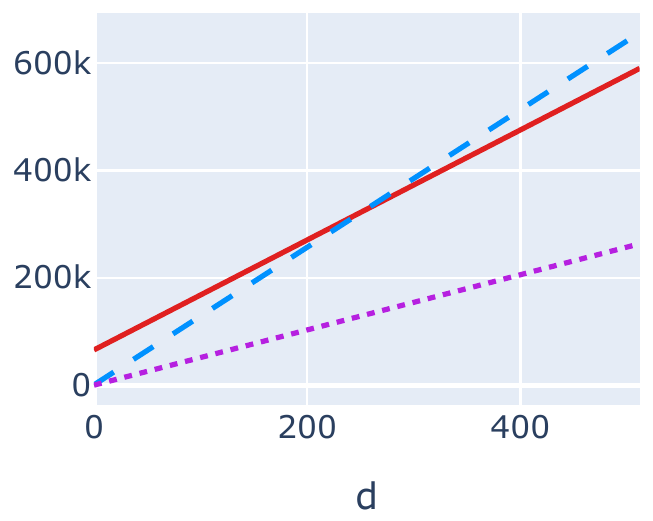}
	    \caption{$n = 256$}
		\label{fig:theory-attn-comp-fixed-n-mem}
	\end{subfigure}
	\caption{\textbf{FLOPs/step and memory footprint} for \mbox{\textcolor{ruby}{\underline{Regular}}}, \textcolor{sapphire}{\dashuline{Continual Retroactive}}, and \textcolor{amethyst}{\dotuline{Continual Single-Output}} Scaled Dot-Product Attention at varying sequence length $n$ and embedding dimension $d$.
	Column (a) has $d$ fixed to 256; Column (b) has $n$ fixed to 256.
	}
	\label{fig:theory-attn-comp}
\end{figure}

%\newpage
\cref{fig:theory-attn-comp} illustrates the scaling of FLOPs and memory footprint with increasing sequence length $n$ and embedding dimension $d$. Here, the Continual Retroactive and Single-Output SDAs spend significantly less FLOPs than the Regular SDA, which scales $\mathcal{O}(n^2)$ as opposed to $\mathcal{O}(nd)$ the continual variants. 
The Continual Single-Output SDA reduces memory footprint for all value combinations, and the Continual Retroactive SDA does so when $n \gtrapprox d$.

%\newpage
% \subsection{Visual depictions of Continual Scaled Dot-Product Attentions}
\subsection{Supplemental visualisations}\label{apx:sup-vis}
\begin{figure}[hb]
    \centering
    \includegraphics[width=0.6\linewidth]{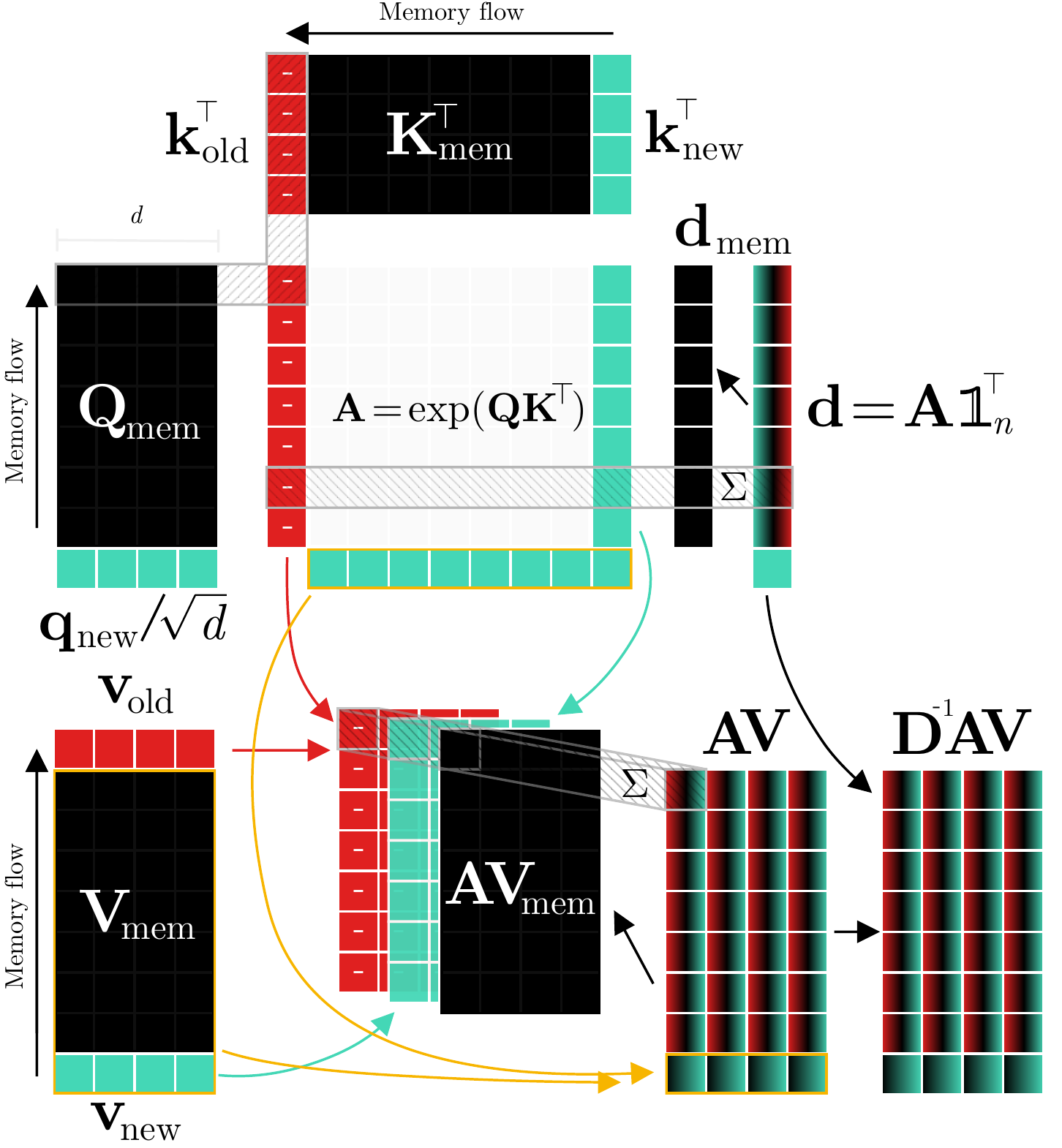}
    \caption{
        \textbf{Continual Retroactive Dot-Product Attention}. 
        The query ($\mQ$), key ($\mK$), and value ($\mV$) matrices are aggregated over time by caching the step vectors \textcolor{teal}{$\vqn$}, \textcolor{teal}{$\vkn$}, and \textcolor{teal}{$\vvn$} in each their FIFO queue (denoted by $\square_\text{mem}$). During each step, only the entries of $\mA$ associated with \textcolor{teal}{$\vqn$}, \textcolor{teal}{$\vkn$} and the oldest $\mK$ step, \textcolor{ruby}{$\vko$} are computed. 
        The diagonal entries of the row-normalisation matrix $\mD$ as well as the $\mA\mV$ can be updated retroactively by subtracting features corresponding to \textcolor{ruby}{$\vko$} and adding features related to \textcolor{teal}{$\vkn$} to the cached outputs of the previous step, $\mD_\text{mem}$ and $\mA\mV_\text{mem}$, respectively.
    }
    \label{fig:co-re-dot-prod-attention}
\end{figure}

\begin{figure}[hb]
    \centering
    \includegraphics[width=0.6\linewidth]{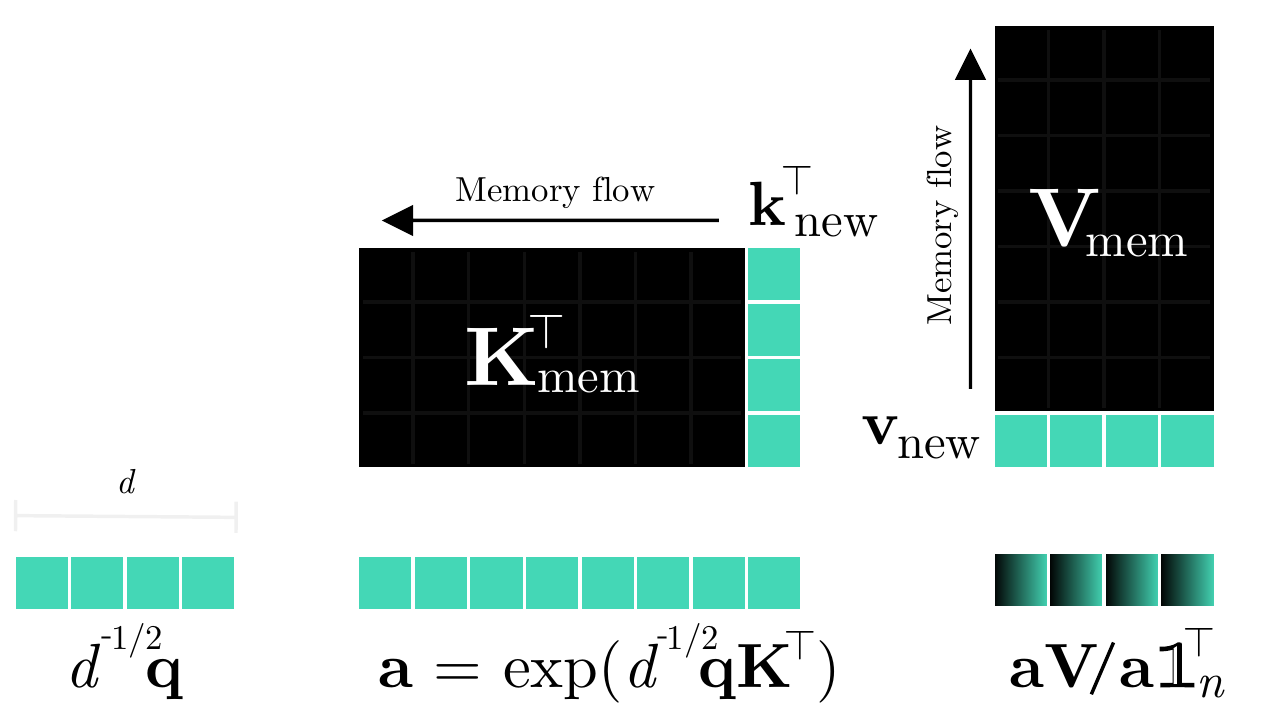}
    \caption{
        \textbf{Continual Single-Output Dot-Product Attention}. 
        The key ($\mK$) and value ($\mV$) matrices are aggregated over time by caching the step vectors $\vkn$ and $\vvn$ in a FIFO queue. During each step, only the attention output associated with $\vq$ is computed.
    }
    \label{fig:cosi-dot-prod-attention}
\end{figure}

For the visually inclined, we supply a complementary graphical depictions of the 
Continual Retroactive SDA corresponding to \crefrange{eq:core-sdpa-d-update}{eq:core-sdpa-dav} in \cref{fig:co-re-dot-prod-attention} and the 
Single-Output SDA in \cref{eq:leading-scaled-dot-product-attention-att} in \cref{fig:cosi-dot-prod-attention}.

% \subsection{Experimental Setup for Audio Classification}

%\newpage
% \subsection{Supplemental visualisations}

% \cref{fig:oad-comparison} supplies as comparison of precision and computational complexity for continual and regular OAD methods.
% A visual comparison of varying class token position for different numbers of Transformer Encoder blocks is given in \cref{fig:oad-cls-ablation}.

% \begin{figure}[t]
%     \centering
%     \includegraphics[width=0.6\linewidth]{figures/oad-comparison.pdf}
%     \caption{
%         \textbf{Visual comparison} of OAD methods on \mbox{THUMOS14} and \mbox{TVSeries} for backbones trained on \mbox{ActivityNet 1.3} and \mbox{Kinetics-400}.
%     }
%     \label{fig:oad-comparison}
% \end{figure}

% \begin{figure}[t]
%     \centering
%     \includegraphics[width=0.6\linewidth]{figures/oad-cls-ablation-epoch2-a-only.pdf}
%     \caption{
%         \textbf{Ablation of \texttt{CLS} token} for varying input positions and transformer encoder blocks (noted as b\#) on THUMOS14 with TSN-Anet features. Black bars show the standard deviation range.
%     }
%     \label{fig:oad-cls-ablation}
% \end{figure}

\begin{figure}[hb]
    \centering
    \includegraphics[width=0.6\linewidth]{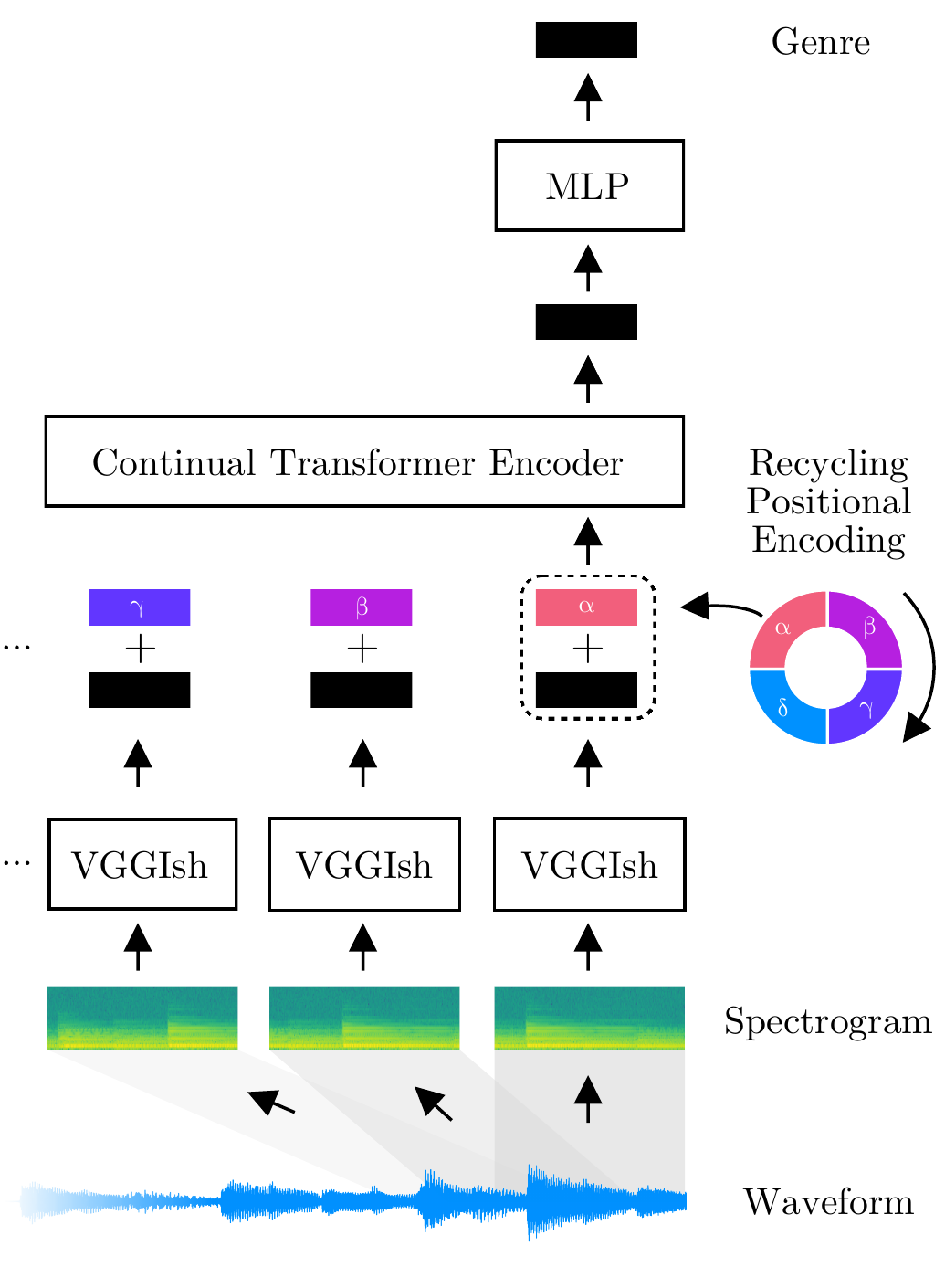}
    \caption{
        \textbf{Audio Classification Architecture}.
    }
    \label{fig:audio_cls}
\end{figure}

A schematic illustration of the Audio Classification experiments architecture is depicted in \cref{fig:audio_cls}.

% \subsection{Training details}\label{apx:training-details}
% In our human activity recognition experiments in \cref{sec:exp-oad}, we follow the setup in \citet{wang2021oadtr} with a batch size of 128, sequence length 64, initial learning rate $10^{-4}$ with a factor ten reduction each epoch, alongside weight decay $10^{-4}$, and dropout with probability $0.1$. We report results using two epochs of training on a Nvidia RTX2080 Ti GPU. 

% For the audio classification experiments in \cref{sec:exp-ac}, we utilize a batch size of 64 and the Adam optimizer~\citep{DBLP:journals/corr/KingmaB14} with an initial learning rate of $10^{-4}$. The learning rate is reduced by a factor of $0.6$ on plateau with a tolerance of two epochs, and an early stopping mechanism, % with a tolerance of five epochs is employed, 
% where a maximum of 100 epochs are allowed. 

\end{document}